\documentclass[lettersize,journal]{IEEEtran}
\usepackage{amsmath,amsfonts}
\usepackage{algorithmic}
\usepackage{algorithm}
\usepackage{array}
\usepackage[caption=false,font=normalsize,labelfont=sf,textfont=sf]{subfig}
\usepackage{textcomp}
\usepackage{stfloats}
\usepackage{url}
\usepackage{orcidlink}
\usepackage{verbatim}
\usepackage{graphicx}
\usepackage{cite}
\usepackage{booktabs}
\usepackage{tabularx} 
\usepackage{adjustbox}
\usepackage{multirow}
\usepackage{makecell}
\newcommand{\cmark}{\checkmark}
\newcommand{\xmark}{\(\times\)}
\newcolumntype{Y}{>{\raggedright\arraybackslash}X}
\makeatletter
\def\endthebibliography{%
	\def\@noitemerr{\@latex@warning{Empty
	`thebibliography' environment}}%
	\endlist
}

\begin{document}

\title{MuS-Polar3D: A Benchmark Dataset for Computational Polarimetric 3D Imaging under Multi-Scattering Conditions}

\author{Puyun~Wang\,\orcidlink{0009-0005-3304-1540} \and Kaimin~Yu\,\orcidlink{0000-0001-7385-6721} \and Huayang~He \and Xianyu~Wu\,\orcidlink{0000-0001-6005-7058}
\thanks{The authors are with the School of Mechanical Engineering and Automation, Fuzhou University, Fuzhou 350108, China (Corresponding Author: Xianyu Wu,  xwu@fzu.edu.cn),
	and with the Research Institute of Highway, Ministry of Transport, Beijing 100088, China.
}
}



\maketitle

\begin{abstract}
Polarization-based underwater 3D imaging exploits polarization cues to suppress background scattering, exhibiting distinct advantages in turbid water. Although data-driven polarization-based underwater 3D reconstruction methods show great potential, existing public datasets lack sufficient diversity in scattering and observation conditions, hindering fair comparisons among different approaches, including single-view and multi-view polarization imaging methods. 
To address this limitation, we construct MuS-Polar3D, a benchmark dataset comprising polarization images of 42 objects captured under seven quantitatively controlled scattering conditions and five viewpoints, together with high-precision 3D models ($\pm$0.05~mm accuracy), normal maps, and foreground masks. The dataset supports multiple vision tasks, including normal estimation, object segmentation, descattering, and 3D reconstruction. 
Inspired by computational imaging, we further decouple underwater 3D reconstruction under scattering into a two-stage pipeline, namely \emph{descattering} followed by \emph{3D reconstruction}, from an imaging-chain perspective. Extensive evaluations using multiple baseline methods under complex scattering conditions demonstrate the effectiveness of the proposed benchmark, achieving a best mean angular error of 15.49$^\circ$. To the best of our knowledge, MuS-Polar3D is the first publicly available benchmark dataset for quantitative turbidity underwater polarization-based 3D imaging, enabling accurate reconstruction and fair algorithm evaluation under controllable scattering conditions. The dataset and code are publicly available at \url{https://github.com/WangPuyun/MuS-Polar3D}.

\end{abstract}


\section{Introduction}
\IEEEPARstart{T}{hree-dimensional} imaging techniques are widely used in applications such as measurement, navigation, and archaeology due to their accurate spatial representation capability. However, in turbid underwater environments, light absorption and scattering caused by suspended particles severely degrade the performance of conventional 3D imaging methods, leading to obscured details, heavy noise, and a significantly reduced effective working range~\cite{Cong2024Survey}. 
Against this backdrop, computational imaging offers a promising direction for overcoming scattering-induced limitations. Unlike traditional computer vision methods that focus solely on post-processing, computational imaging emphasizes modeling and optimizing the entire imaging chain. By encoding information across different imaging stages, it enables the joint design of the imaging system and reconstruction algorithms, thereby breaking conventional performance bottlenecks~\cite{xiang2024computational}. 
Inspired by this principle, we revisit the 3D imaging pipeline in underwater scattering media and decouple the overall process into two stages: descattering followed by 3D reconstruction. Under this framework, among various available optical priors, light polarization exhibits distinct advantages in both stages. Consequently, this work leverages polarization information to exploit its inherent capability in scattering suppression and structural inference, aiming to improve 3D reconstruction accuracy in complex underwater environments where conventional vision-based methods often fail.

\begin{figure}[!t]
	\centering
	\includegraphics[width=0.49\textwidth]{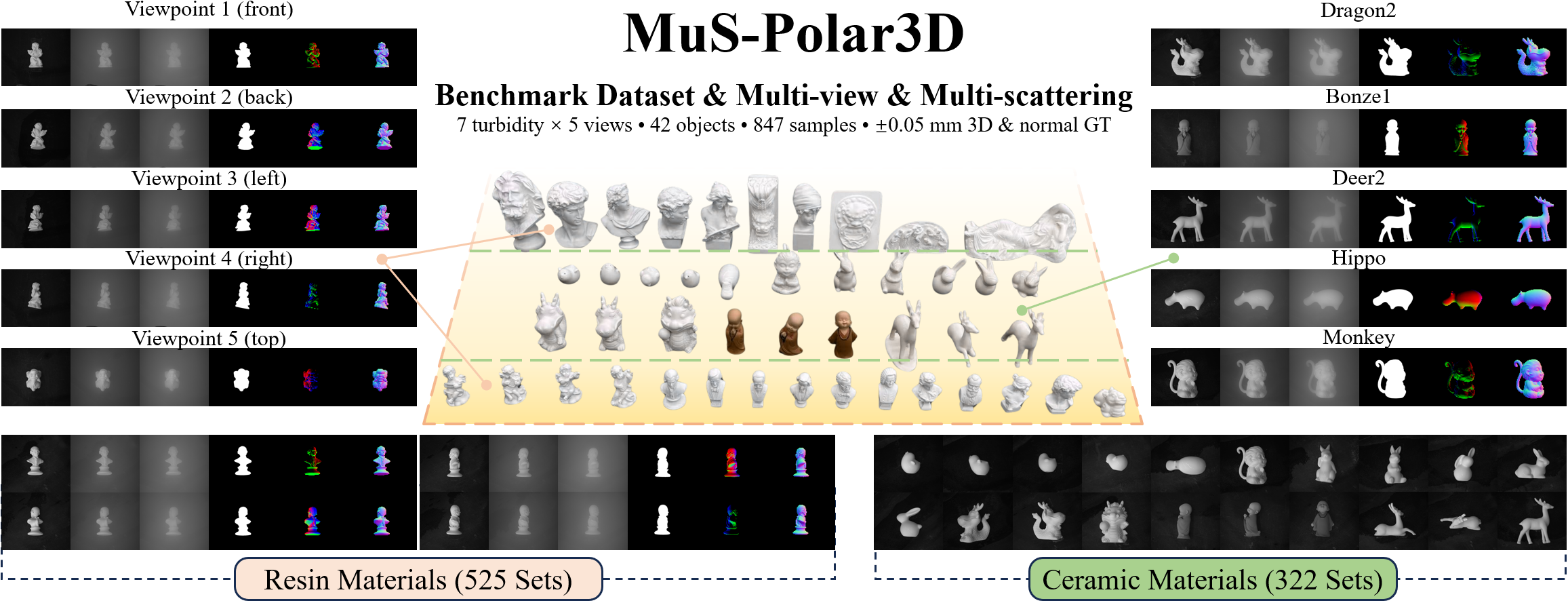}
	\caption{\textbf{Overview of MuS-Polar3D.} The MuS-Polar3D dataset comprises 42 distinct objects and over 800 well-processed samples. These include both ceramic and resin objects, as well as high-texture and low-texture targets, ensuring substantial diversity across the dataset.}
	\label{fig:Overview of MuS-Polar3D}
\end{figure}

Since Ba et al.~\cite{ba2020deep} first introduced deep learning into polarization-based 3D imaging in 2020, this line of research has advanced rapidly~\cite{lyu2024sfpuel,tiwari2024ss}. Wu et al.~\cite{wu2025deep} extended polarization-based 3D reconstruction to underwater scenarios. By developing a deep learning framework based on U\textsuperscript{2}-Net~\cite{qin2020u2}, they demonstrated high reconstruction accuracy and strong robustness even in complex underwater environments.

Despite these advances, the scarcity of high-quality polarization underwater datasets remains a major bottleneck for developing new methods and conducting systematic evaluations. Specifically, most existing datasets~\cite{ba2020deep,lyu2024sfpuel,collins2022abo,lei2022shape,shao2023transparent,reizenstein2021common} provide only single-view or limited multi-view observations, making it difficult to fairly compare polarization-based 3D imaging methods with recent multi-view techniques, such as neural implicit signed distance field (SDF)-based approaches~\cite{chen2024uw}, multi-view stereo (MVS)~\cite{luo2024lnmvsnet}, and structure-from-motion (SfM). This limitation hinders the comprehensive evaluation of polarization-based methods within the broader 3D vision landscape. 
Moreover, these datasets are typically collected under standard imaging conditions and fail to capture the diversity of real underwater environments in terms of turbidity and scattering strength, which in turn restricts the generalization ability of algorithms across different water conditions.

In summary, existing public datasets rarely provide both multi-view observations and diverse turbidity conditions, making them insufficient for algorithm development and systematic benchmarking. To address this gap, we present MuS-Polar3D, a comprehensive underwater polarization benchmark dataset comprising 42 objects captured under seven scattering conditions and up to five viewpoints. MuS-Polar3D offers a more complete and challenging experimental platform for polarization-based 3D imaging research.

The main contributions of this work are summarized as:

\begin{itemize}
	\item MuS-Polar3D is constructed as a benchmark dataset for passive underwater polarization imaging, supporting multiple 3D reconstruction paradigms. The dataset provides multi-view polarization images under diverse turbidity conditions, together with high-precision 3D models, ground-truth surface normals, foreground masks, and detailed sample annotations. It supports both single-view polarization-based shape-from-polarization (SfP) reconstruction and serves as a unified evaluation benchmark for general multi-view 3D imaging methods, such as SDF-based approaches and SfM/MVS pipelines.
	
	\item The comparability gap between SDF-based multi-view methods and polarization-based SfP approaches in underwater 3D imaging is addressed. By analyzing the intrinsic differences between neural implicit SDF modeling and physics-driven polarization modeling, a unified multi-view data acquisition and experimental protocol is established, enabling fair and consistent cross-paradigm comparisons.
	
	\item A two-stage 3D imaging framework for scattering media is introduced, consisting of descattering followed by 3D reconstruction. Starting from the physical degradation mechanisms of passive underwater imaging, the variation of normal estimation accuracy under different turbidity levels is systematically analyzed, experimentally revealing the relationship between scattering severity and 3D reconstruction performance.
	
	\item A complete benchmarking and analysis framework for underwater polarization-based 3D imaging is established. Dedicated qualitative and quantitative evaluation metrics are designed for both the descattering and 3D reconstruction stages, and representative methods, including DeepSfP and SfPW, are integrated into a unified evaluation pipeline, reducing experimental barriers and improving reproducibility and fairness in method comparison.
\end{itemize}

\section{Related Work}

\subsection{Imaging in Turbid Water}

Imaging in scattering media is highly challenging due to the spatial non-uniformity and temporal dynamics of the medium, which lead to complex and difficult-to-model light propagation. Such heterogeneous scattering significantly degrades image contrast and resolution, making underwater imaging in turbid water particularly difficult~\cite{zhu2018recovering}. 
Benefiting from strong nonlinear representation capabilities, deep learning methods can automatically learn implicit relationships between target structures and scattering-induced degradations from data, enabling effective underwater descattering. In recent years, extensive efforts have been devoted to mitigating scattering effects using deep learning-based approaches. Liu et al.~\cite{liu2024learning} proposed a cross-domain deep learning framework for real-time underwater descattering. By systematically analyzing the physical interactions between light and object surfaces, they innovatively employed an electronic ink display to simulate optical properties of targets for training data acquisition, achieving high image quality and robustness while maintaining real-time performance. 
Li et al.~\cite{li2024underwater} introduced a polarization-based descattering model that incorporates adaptive particle swarm optimization and dual-domain contrast enhancement, effectively suppressing backscattering in real turbid water environments. Fu et al.~\cite{fu2022uncertainty} addressed ambiguities among multiple enhancement results by leveraging probabilistic distribution estimation and consensus strategies, achieving state-of-the-art performance on large-scale real-world datasets. Collectively, these methods substantially improve the visual clarity of underwater images, providing a solid foundation for high-quality underwater 3D reconstruction.

\subsection{Polarization-Based 3D Imaging}

Compared with the random vibration directions of the electric field in natural light, polarized light exhibits regular and consistent electric field oscillations. As a result, polarization-based 3D reconstruction often yields more stable and accurate results than conventional intensity-based imaging. Polarization introduces additional physically meaningful dimensions to 3D reconstruction, providing critical cues such as surface normals and material properties, thereby complementing traditional geometry-based or photometric reconstruction methods.

For turbid underwater environments, Yang et al.~\cite{yang2024high} recovered clear polarization information using a deep descattering network and reconstructed high-quality 3D geometry based on an improved physical model. Ding et al.~\cite{ding2021polarimetric} incorporated polarization phase into a Helmholtz stereopsis framework through a novel reciprocal optical configuration, improving reconstruction performance on complex material surfaces.

Deep learning methods have further demonstrated strong potential in this field. Ba et al.~\cite{ba2020deep} first integrated polarization physics with deep learning-based 3D shape reconstruction using a hybrid training strategy that combines physical priors and data-driven learning, reducing the normal estimation error on real datasets from approximately 40$^\circ$ with traditional physical methods to about 18$^\circ$. Lei et al.~\cite{lei2022shape} addressed outdoor scenarios by introducing camera viewpoint encoding to handle perspective effects, and further enhanced the applicability and robustness of shape-from-polarization through efficient polarization representations and multi-head attention mechanisms. Lyu et al.~\cite{lyu2024sfpuel} investigated the impact of unknown illumination on polarization-based 3D reconstruction and proposed a method that integrates polarization cues, photometric stereo priors, and global attention modeling, enabling high-quality normal reconstruction from a single polarization image under unknown lighting conditions while automatically distinguishing between metallic and dielectric materials.

\begin{table*}[!htbp]
	\caption{Summary of polarization/scattering-related datasets.
		Abbrev.: ENH=image enhancement, SNE=surface normal estimation,
		DET=object detection (mask), MAT=material property estimation, 3D=3D reconstruction.}
	\label{tab:dataset-summary}
	\centering
	\setlength{\tabcolsep}{3pt}
	\renewcommand{\arraystretch}{1.3}
	
	\begin{adjustbox}{width=\textwidth,center}
		\begin{tabularx}{\textheight}{@{}l c c l cccc c c Y c @{}}
			\toprule
			Dataset & Medium & Concentration Variation & Modality &
			\multicolumn{4}{c}{Ground Truth} &
			Multi-view & Material Diversity & Application Focus & Year \\
			\cmidrule(lr){5-8}
			&  &  &  & Clean & 3D Model & Normal & Mask &  &  &  &   \\
			\midrule
			SfPUEL~\cite{lyu2024sfpuel}      & Air                   & \xmark & Pol.\ RGB                     & \xmark & \xmark & \cmark & \cmark & \xmark & \cmark & SNE                & 2024  \\
			DeepSfP~\cite{ba2020deep}     & Air                   & \xmark & Pol.\ grayscale               & \xmark & \xmark & \cmark & \cmark & \xmark & \cmark & SNE                & 2020 \\
			TransSfP~\cite{shao2023transparent}    & Air                   & \xmark & Pol.\ grayscale               & \xmark & \xmark & \cmark & \cmark & \xmark & \xmark & SNE                & 2023 \\
			SfPW~\cite{lei2022shape}         & Air & \xmark & Pol.\ RGB                     & \xmark & \xmark & \cmark & \xmark & \xmark & \cmark & SNE                & 2022 \\
			MuS-Polar3D & Controlled emulsion & \cmark & Pol.\ grayscale + 3D mesh models & \cmark & \cmark & \cmark & \cmark & \cmark & \cmark & ENH; SNE; DET; MAT; 3D & 2025 \\
			\bottomrule
		\end{tabularx}
	\end{adjustbox}
\end{table*}

\subsection{Public Datasets for Polarization-Based 3D Imaging}

A summary of public polarization datasets for 3D reconstruction is provided in Table~\ref{tab:dataset-summary}. Overall, most existing datasets are collected in controlled indoor or outdoor air environments, with limited consideration of multi-view observations and little systematic coverage of diverse scattering conditions.

Specifically, the SfPUEL dataset~\cite{lyu2024sfpuel} focuses on polarization imaging under varying environmental illumination conditions. By constructing large-scale synthetic and real datasets, it addresses issues such as limited data scale and the scarcity of real metallic objects in existing benchmarks. DeepSfP~\cite{ba2020deep} introduced the first real object-level shape-from-polarization dataset, capturing polarization images of 25 objects under four different poses across three illumination conditions, including indoor lighting, outdoor overcast, and outdoor direct sunlight.
TransSfP~\cite{shao2023transparent} constructed the first polarization dataset dedicated to transparent objects, consisting of both synthetic and real data. The real data were captured using a polarization camera and a 3D scanner and subsequently aligned in Unity, while the synthetic data were generated using the physically based rendering engine Mitsuba~\cite{jakob2022dr} with full-polarization rendering to simulate realistic illumination, transmission, and reflection processes. The SfPW dataset~\cite{lei2022shape} targets scene-level 3D reconstruction by acquiring raw polarization images with a polarization camera and generating corresponding ground-truth surface normals using a depth camera. It contains 522 real-world scenes covering diverse material types, illumination conditions, and depth ranges.

\section{Dataset Construction}

\begin{figure}[!htbp]
	\centering
	\includegraphics[width=0.7\linewidth]{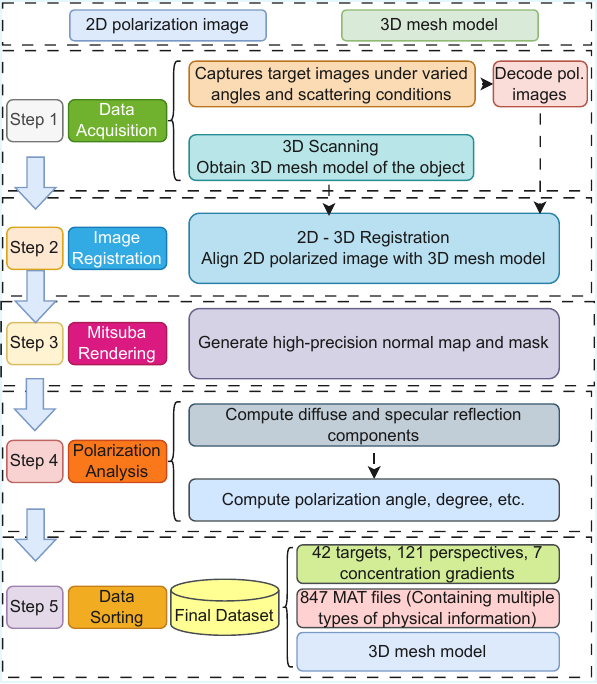}
	
	\caption{Pipeline of the MuS-Polar3D dataset construction.}
	\label{fig:dataset_pipeline}
\end{figure}

Previous studies have shown that emulsions can simulate the optical scattering properties of seawater~\cite{michels2008optical}. Based on this, a complete pipeline for data acquisition, processing, and organization is established, as illustrated in Fig.~\ref{fig:dataset_pipeline} .

\subsection{Acquisition System and Data Collection}
\subsubsection{Polarization Image Acquisition}

\begin{figure*}[!htb]
	\centering
	\begin{minipage}[t]{0.30\linewidth}
		\centering
		\includegraphics[width=\linewidth]{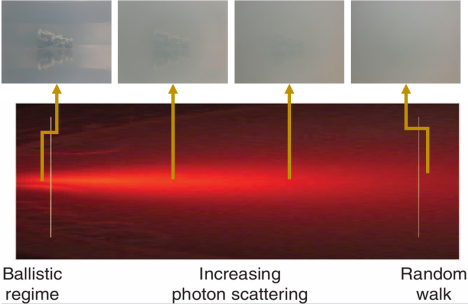} 
		\vspace{2pt}\scriptsize (a) Classification of scattering regions.
	\end{minipage}
	\begin{minipage}[t]{0.33\linewidth}
		\centering
		\includegraphics[width=\linewidth]{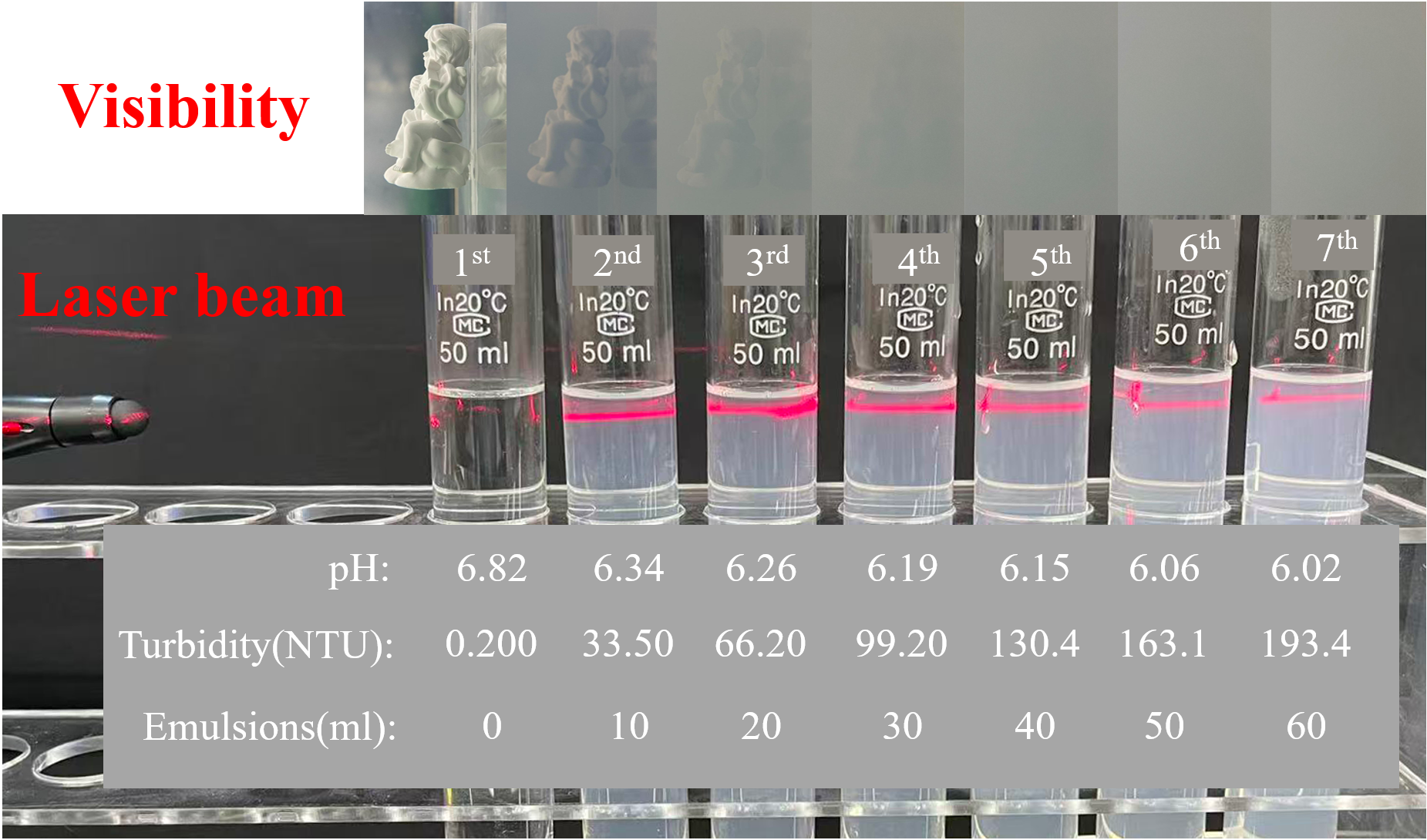} 
		\vspace{2pt}\scriptsize (b) Quantification of scattering degree
	\end{minipage}
	\begin{minipage}[t]{0.16\linewidth}
		\centering
		\includegraphics[width=\linewidth]{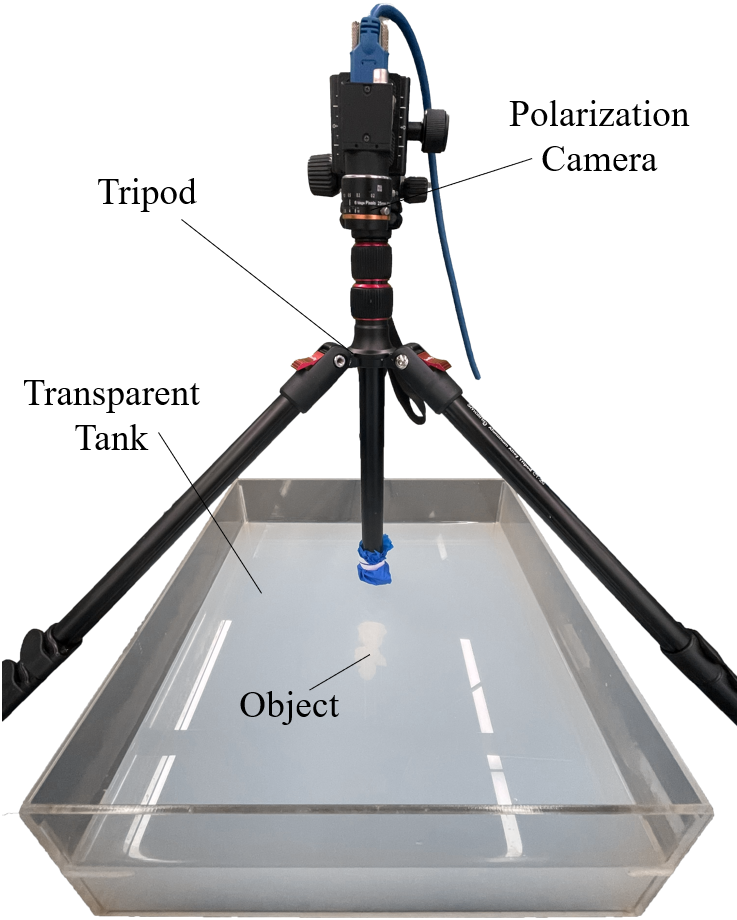} 
		\vspace{2pt}\scriptsize (c) Imaging setup
	\end{minipage}
	\caption{Schematic diagram  of the experimental setup for simulating different scattering intensities and the underwater passive polarization imaging model.}
	\label{fig:acquisition_and_scattering_conditions}
\end{figure*}

The scattering process experienced by light propagation in underwater environments is illustrated in Fig.~\ref{fig:acquisition_and_scattering_conditions}(a). In general, the degree of scattering increases rapidly with propagation distance or water turbidity. According to scattering intensity, the imaging region can be divided into three typical regimes: the ballistic regime, the increasing photon scattering regime, and the random walk regime~\cite{ntziachristos2010going}. 
In the ballistic regime, scattering effects are relatively weak, and clear observations can be obtained through direct imaging. In the increasing photon scattering regime, the intensity distribution of scattered light approximately follows the variation described in Eq.~\ref{eq:scattering_model}~\cite{treibitz2008active}. In contrast, in the random walk regime, light undergoes strong multiple scattering and largely loses its directionality, making it difficult to acquire effective target information through direct imaging.

\begin{equation}
	\begin{aligned}
		I_{\mathrm{scat}}(x,y)
		&=
		\int_{R_{\mathrm{cam}}=0}^{R_{\mathrm{cam}}(x_{\mathrm{obj}})}
		b\!\left[\theta(z)\right]\,
		I_{\mathrm{source}}(z) \\
		&\quad \times
		\exp\!\left(-c\,R_{\mathrm{cam}}(z)\right)
		\,\mathrm{d}R_{\mathrm{cam}} ,
	\end{aligned}
	\label{eq:scattering_model}
\end{equation}
where $R_{\mathrm{cam}}(z)$ denotes the distance between an arbitrary scene point and the detector, $I_{\mathrm{source}}(z)$ represents the irradiance of the light source, and $\theta \in [0,\pi]$ is the scattering angle of the light wave within the scattering medium. The coefficient $b$ denotes the scattering coefficient of the medium, characterizing the backscattering strength at a given angle $\theta$, and $c$ is the attenuation coefficient of the medium.

Based on the above model, emulsions are employed to simulate different levels of scattering degradation in an underwater passive polarization imaging model~\cite{liu2019polarization}, enabling controllable and reproducible experimental conditions. During data acquisition, emulsions are gradually added to water from 0~ml to 60~ml with a step size of 10~ml, resulting in seven scattering levels ranging from clear to highly turbid conditions. The turbidity values corresponding to each scattering level are accurately calibrated using a turbidity meter, as shown in Fig.~\ref{fig:acquisition_and_scattering_conditions}(b).
The polarization camera is positioned outside a transparent water tank, while a black matte material is used as the imaging background to suppress unwanted background reflections. The detailed setup is illustrated in Fig.~\ref{fig:acquisition_and_scattering_conditions}(c). All polarization images are captured under constant indoor illumination and controlled medium conditions to simulate underwater passive polarization imaging scenarios~\cite{liu2019polarization}. The polarization camera used in this work is a FLIR BFS-U3-51S5P-C, which simultaneously captures four polarization states at 0$^\circ$, 45$^\circ$, 90$^\circ$, and 135$^\circ$ within a single exposure~\cite{yamazaki2016four}.

\subsubsection{3D Mesh Acquisition}

All target objects are digitized using a high-precision structured-light 3D scanner from the Shining3D EinScan series. Based on structured-light measurement principles, the scanner acquires surface geometry from multiple viewpoints, enabling non-contact and high-accuracy reconstruction of object shapes. During scanning, point clouds captured from different viewpoints are registered and fused, followed by noise removal and mesh reconstruction to generate complete and geometrically accurate 3D mesh models. The overall accuracy of the reconstructed models reaches $\pm$0.05~mm, allowing faithful representation of object dimensions and surface details.

\begin{figure}[!t]
	\centering
	\includegraphics[width=0.3\textwidth]{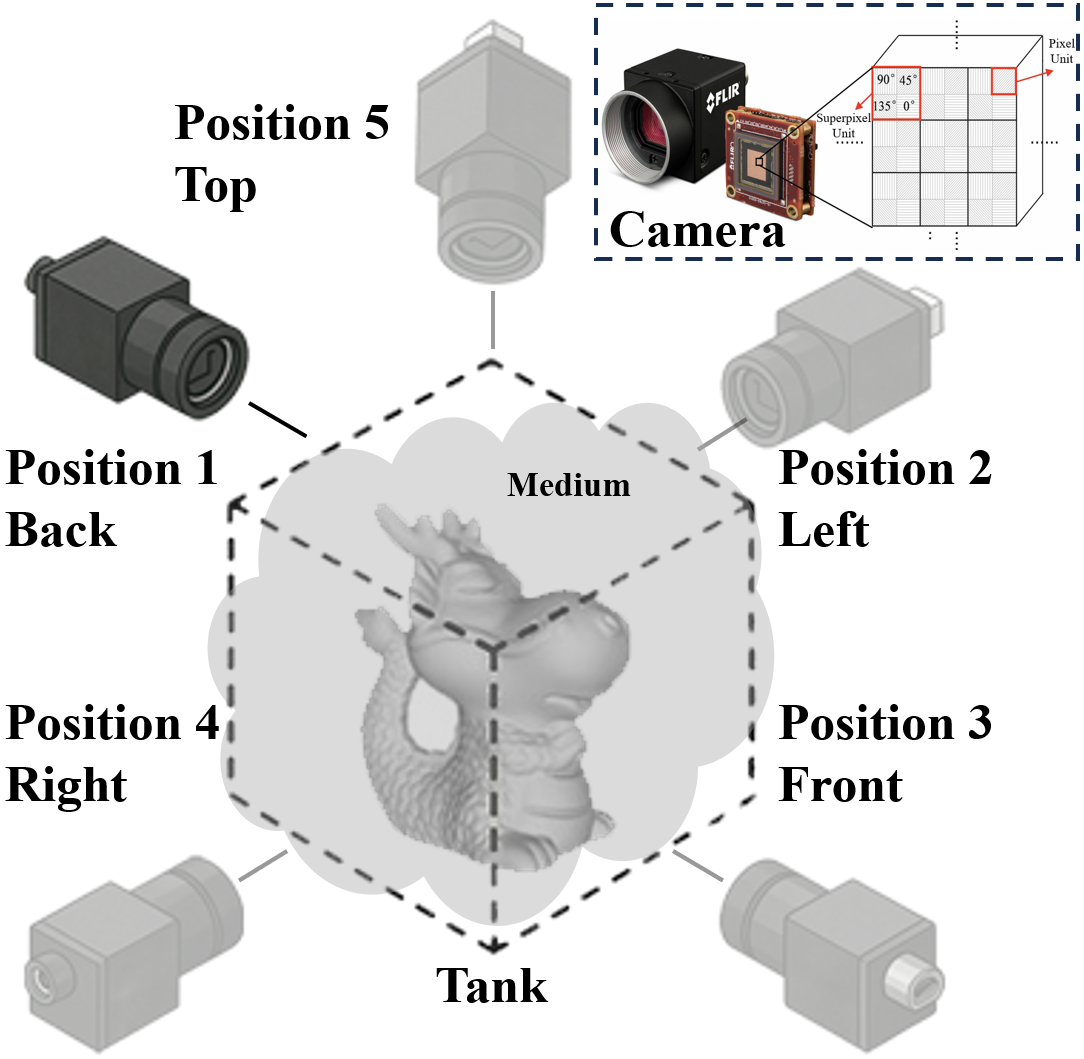}
	\caption{Schematic illustration of the multi-view imaging.}
	\label{fig:multiview_setup}
\end{figure}

\subsubsection{Objects and Viewpoints}
\label{sec:Objects and Viewpoints}
To ensure dataset diversity, a variety of target objects are selected, including ceramic crafts and resin sculptures with complex 3D geometries. The surface textures of these objects span a wide range, from richly textured to relatively dull appearances.

\begin{figure*}[!ht]
	\centering
	\includegraphics[width=0.9\linewidth]{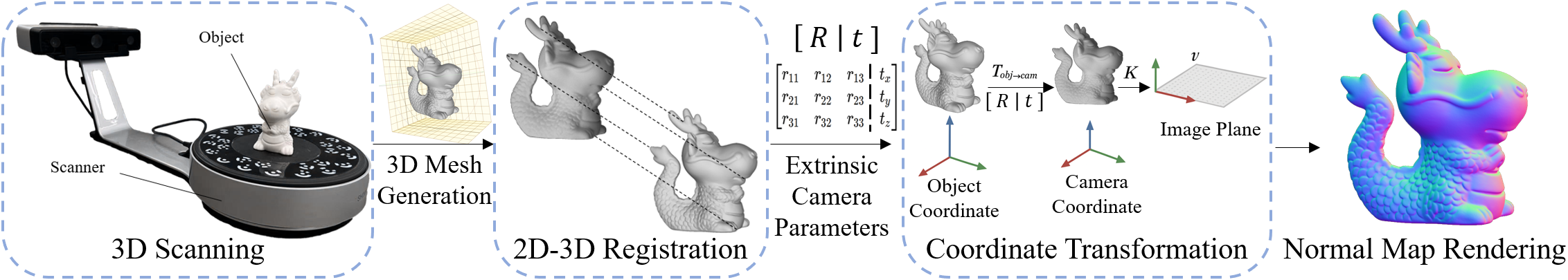}
	\caption{Workflow of the normal map rendering based on 3D scanning and geometric registration.}
	\label{fig:normal_map_generation}
\end{figure*}

Material properties are inherent characteristics of each target object and constitute a key physical factor affecting the stability and accuracy of 3D reconstruction. The materials included in this dataset mainly consist of resin sculptures, glazed ceramics, and unglazed ceramics, with representative examples shown in Fig.~\ref{fig:classification_results}(c).

From an optical perspective, resin sculptures are dominated by diffuse reflection with relatively weak specular components. Glazed ceramics exhibit pronounced specular reflections, leading to significant inconsistencies in highlight regions across multiple viewpoints. In contrast, unglazed ceramics contain both diffuse and specular reflection components, presenting a mixed reflection behavior.

Differences in the proportion and spatial distribution of reflection components among these materials directly lead to complex degradations in underwater images, including intensity inconsistencies, view-dependent specular highlights, and variations in polarization states. Such material-induced imaging variations make it difficult for conventional 3D reconstruction methods~\cite{mahmoud2012direct,miyazaki2004transparent,smith2016linear}, which typically assume purely specular or purely diffuse surface reflectance, to achieve stable modeling, often resulting in geometric errors or reconstruction failures on objects with complex materials. By systematically incorporating objects with representative reflection mechanisms at the data level, the proposed dataset provides the necessary conditions for analyzing the impact of material physical properties on underwater 3D reconstruction performance.

During polarization image acquisition, each object is placed at the center of the water tank and sequentially rotated to five viewpoints, including front, back, left, right, and top views, to obtain multi-view polarization observations, as illustrated in Fig.~\ref{fig:multiview_setup}. The statistics of object counts and proportions for the three material categories are further reported in Section~\ref{sec:data_analysis} .

\subsection{Data Items and Annotation Generation}

After raw data acquisition, each sample consists of four polarization-state images and the corresponding 3D mesh model. A series of processing steps is then applied to generate complete annotations, forming the final dataset.

\subsubsection{Registration and Ground-Truth Generation}

To establish accurate correspondence between polarization images and 3D geometric models, a semi-automatic registration strategy is adopted using the built-in registration algorithms~\cite{corsini2009image} in MeshLab~\cite{cignoni2008meshlab}. Specifically, an automatic registration procedure is first applied to achieve coarse alignment and obtain reasonable camera pose estimates. This is followed by interactive manual refinement to further improve registration accuracy, resulting in the final determination of camera intrinsics and extrinsics for each viewpoint.

With camera parameters obtained, the 3D models are rendered using the physically based rendering engine Mitsuba~3~\cite{jakob2022dr} to generate high-quality ground-truth surface normal maps and foreground masks that are strictly aligned with the real acquisition viewpoints. This pipeline ensures consistency between 2D polarization observations and 3D geometric ground truth, providing a reliable geometric and annotation foundation for subsequent polarization-based descattering, normal estimation, and 3D reconstruction evaluation.

\subsubsection{Polarization Prior Computation}

A set of polarization-related physical priors is additionally computed and stored to support subsequent studies. This subsection details the computation of the polarization quantities used in this work.

To characterize the polarization state of each target, the Stokes vector is first computed from four polarization-state images, namely $I_{0^\circ}$, $I_{45^\circ}$, $I_{90^\circ}$, and $I_{135^\circ}$. The Stokes vector jointly encodes the total intensity of the light field and its linear polarization components~\cite{kuehn2015student}, and is computed as
\begin{equation}
	\mathbf{S} =
	\begin{bmatrix}
		S_0 \\
		S_1 \\
		S_2
	\end{bmatrix}
	=
	\begin{bmatrix}
		I_{0^\circ} + I_{90^\circ} \\
		I_{0^\circ} - I_{90^\circ} \\
		I_{45^\circ} - I_{135^\circ}
	\end{bmatrix}.
	\label{eq:stokes}
\end{equation}

Here, $S_0$ denotes the total irradiance without considering polarization effects, $S_1$ represents the intensity difference between horizontal and vertical polarization components, and $S_2$ corresponds to the intensity difference between the $+45^\circ$ and $-45^\circ$ polarization directions. Based on the Stokes vector, the angle of polarization (AoP) and the degree of polarization (DoP) can be further computed as follows:

\begin{equation}
	\mathrm{AoP} = \frac{1}{2}\arctan\!\left(\frac{S_2}{S_1}\right),
	\label{eq:aop}
\end{equation}
\begin{equation}
	\mathrm{DoP} = \frac{\sqrt{S_1^2 + S_2^2}}{S_0}.
	\label{eq:dop}
\end{equation}

The relationship between the angle of polarization (AoP) and the surface normal depends on whether surface reflection is dominated by diffuse or specular components. Accordingly, the azimuthal component of the surface normal, denoted as $\phi$, can be determined as

\begin{equation}
	AoP =
	\begin{cases}
		\phi, & \text{for diffuse reflection},\\[4pt]
		\phi + \tfrac{\pi}{2}, & \text{for specular reflection.}
	\end{cases}
\end{equation}

Under the assumption that surface reflection is dominated by diffuse components, the DoP computed from the Stokes parameters can be approximated by the theoretical diffuse polarization degree $\rho_d$. Accordingly, the zenith angle $\theta$ of the surface normal can be inferred from the measured DoP using
\begin{equation}
	\rho_d =
	\frac{\left(n-\frac{1}{n}\right)^2 \sin^2 \theta}
	{2(1+n^2) - \left(n+\frac{1}{n}\right)^2 \sin^2 \theta + 4 \cos \theta \sqrt{n^2-\sin^2 \theta}},
	\label{eq:rho_d}
\end{equation}
where $n$ denotes the refractive index of the medium.

When surface reflection is dominated by specular components, the corresponding polarization degree $\rho_s$ is also determined by the zenith angle $\theta$ and the refractive index $n$, and can be expressed as
\begin{equation}
	\rho_s =
	\frac{2 \sin \theta \tan \theta \sqrt{n^2-\sin^2 \theta}}
	{n^2 - 2 \sin^2 \theta + \tan^2 \theta}.
	\label{eq:rho_s}
\end{equation}

Within a certain angular range, both the diffuse model $\rho_d(\theta)$ and the specular model $\rho_s(\theta)$ exhibit monotonic behavior. However, unlike the diffuse case, the specular model may yield two possible zenith angle solutions for a given polarization degree.

Given a known refractive index $n$, the surface normal vector $\mathbf{n}$ can be further expressed as
\begin{equation}
	\mathbf{n} = [n_x, n_y, n_z]^{\mathsf{T}}
	= [\cos \theta \cos \phi,\; \cos \theta \sin \phi,\; \sin \theta]^{\mathsf{T}}.
	\label{eq:normal}
\end{equation}

\section{Data Analysis}
\label{sec:data_analysis}

After the dataset construction is completed, a systematic analysis is conducted. Specifically, multiple attributes are annotated and statistically analyzed for each sample, including object size, texture richness, and material properties. To quantify the size characteristics of each sample, the number of valid pixels in the mask image is computed as
\begin{equation}
	N_{\mathrm{valid}} = \sum_{x,y} \mathbb{1}\!\left[M(x,y)=1 \wedge V(x,y)=1\right],
	\label{eq:nvalid}
\end{equation}
where $M(x,y)$ denotes the foreground object mask, and $V(x,y)$ indicates whether a valid surface normal exists at pixel $(x,y)$. Based on the image resolution $H \times W$, the effective pixel ratio of each sample is further defined as
\begin{equation}
	R_{\mathrm{eff}} = \frac{N_{\mathrm{valid}}}{H \times W}.
	\label{eq:reff}
\end{equation}

This metric reflects the spatial coverage of each target object in the image, thereby providing a quantitative characterization of object size.

To further describe texture characteristics, the Local Binary Pattern (LBP) operator~\cite{ojala2002multiresolution} is adopted to quantify texture richness through statistical analysis of LBP features, which is defined as
\begin{equation}
	\mathrm{LBP}(x,y)
	=
	\sum_{p=0}^{P-1} s\!\left(I_p - I_c\right) 2^p,
	\quad
	s(z)=
	\begin{cases}
		1, & z \ge 0, \\
		0, & z < 0,
	\end{cases}
	\label{eq:lbp}
\end{equation}
where $I_c$ denotes the intensity of the center pixel, and $I_p$ represents the intensity of the $p$-th neighboring pixel within a circular neighborhood of radius $r$.
The object-level texture statistic is computed as the average LBP value over all valid pixels, given by
\begin{equation}
	\overline{\mathrm{LBP}} =
	\frac{1}{N_{\mathrm{valid}}}
	\sum_{x,y} \mathrm{LBP}(x,y).
	\label{eq:mean_lbp}
\end{equation}

After quantifying object size using $R_{\mathrm{eff}}$ and characterizing texture complexity with $\overline{\mathrm{LBP}}$, a K-means clustering algorithm~\cite{pedregosa2011scikit} is applied to hierarchically partition the dataset into three levels. The clustering objective is formulated as
\begin{equation}
	\min_{\{C_k\}_{k=1}^{K}} 
	\sum_{k=1}^{K} 
	\sum_{i \in C_k} 
	\left\|
	\mathbf{x}_i - \boldsymbol{\mu}_k
	\right\|_2^2,
	\quad
	\mathbf{x}_i = 
	\begin{bmatrix}
		R_{\mathrm{eff}}^{(i)} \\
		\overline{\mathrm{LBP}}^{(i)}
	\end{bmatrix},
	\quad
	K = 3,
\end{equation}
where each object $i$ is represented by a two-dimensional feature vector consisting of its effective pixel ratio and LBP-based texture measure.

\begin{figure*}[!ht]
	\centering
	
	\begin{minipage}[t]{0.31\linewidth}
		\centering
		\includegraphics[width=\linewidth]{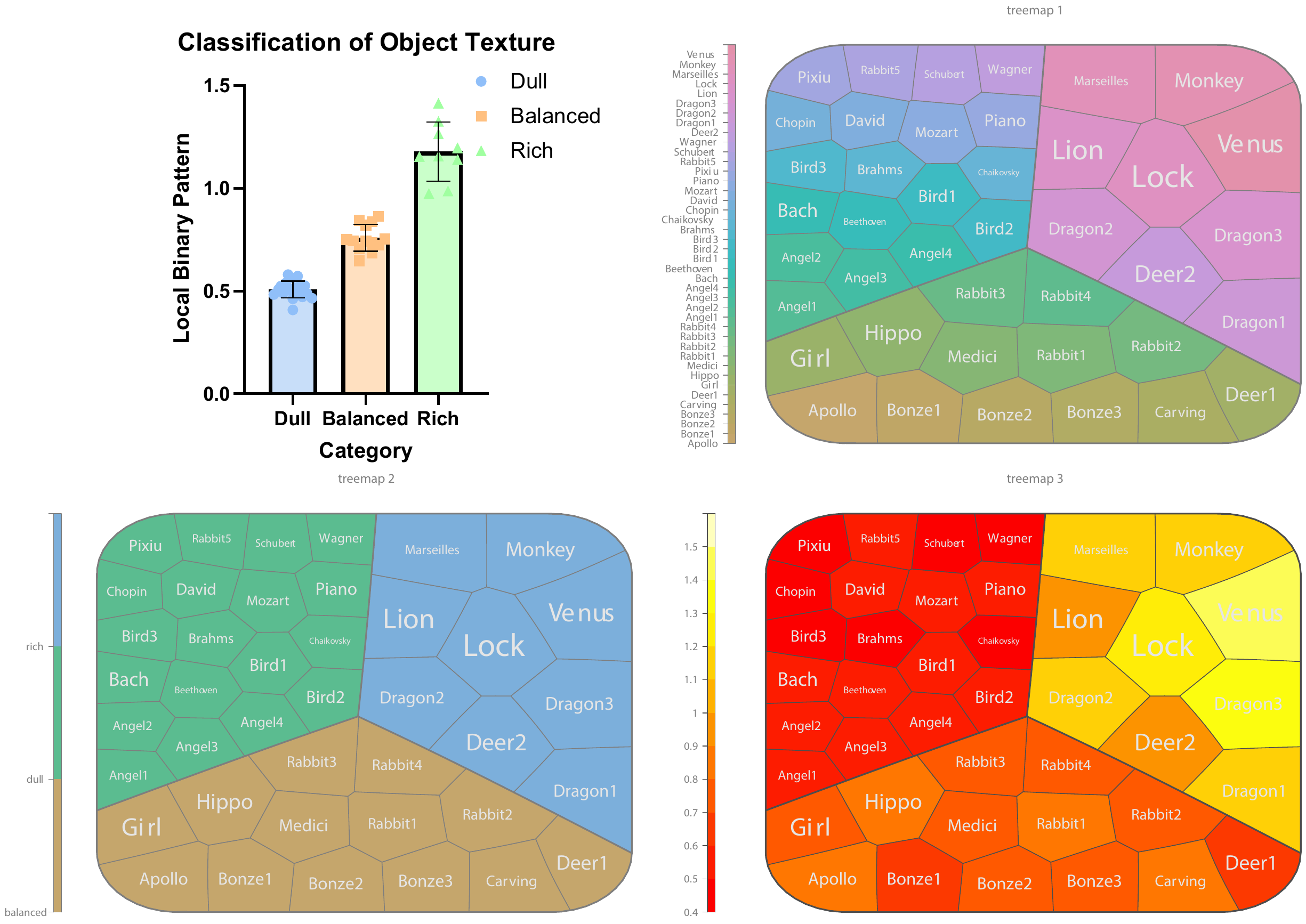} 
		\vspace{2pt}\scriptsize (a) Texture classification
	\end{minipage}
	\begin{minipage}[t]{0.31\linewidth}
		\centering
		\includegraphics[width=\linewidth]{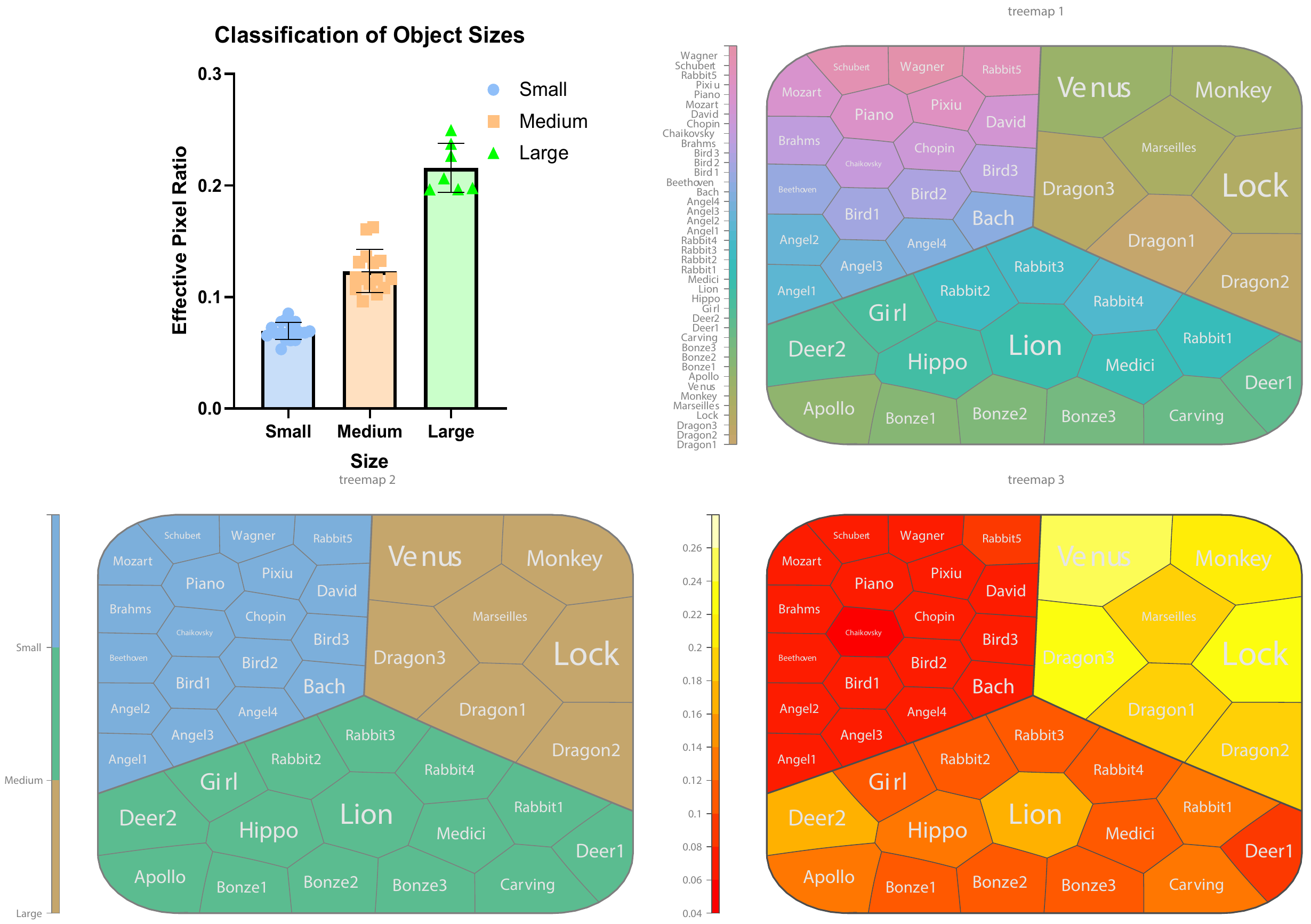} 
		\vspace{2pt}\scriptsize (b) Size classification
	\end{minipage}\hfill
	\begin{minipage}[t]{0.37\linewidth}
		\centering
		\includegraphics[width=\linewidth]{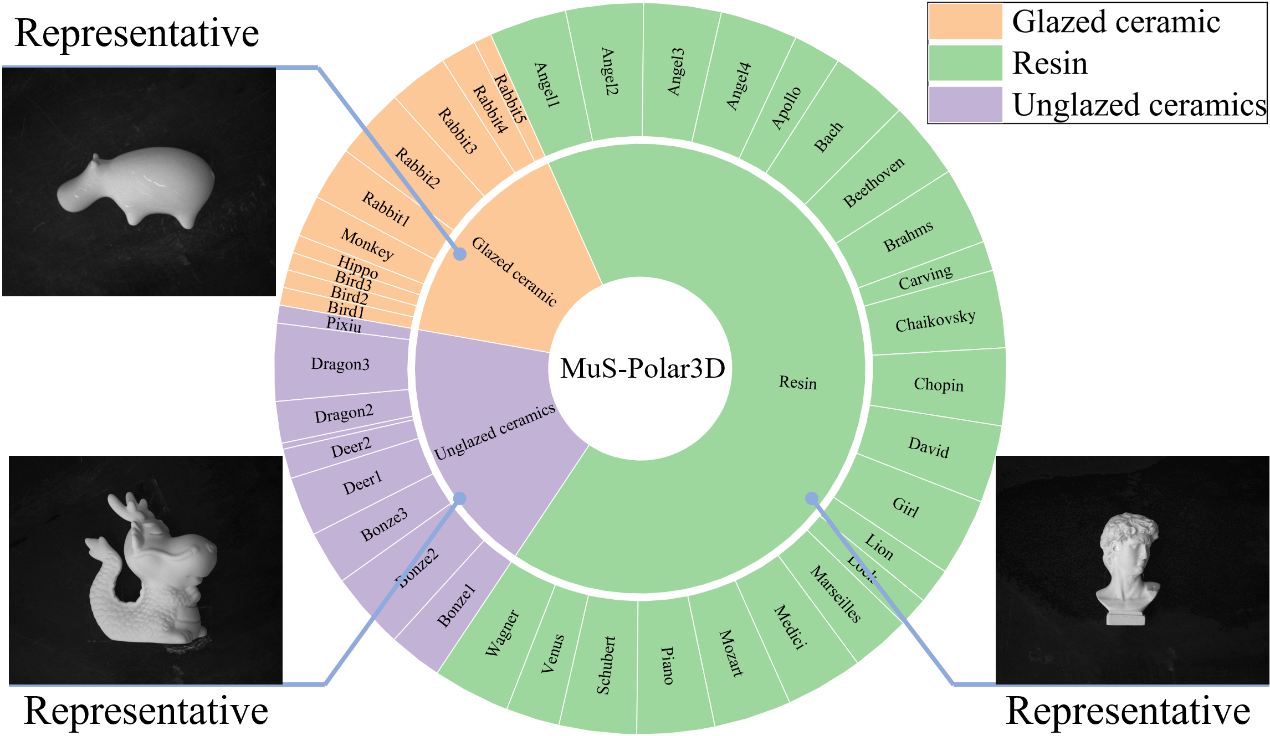} 
		\vspace{2pt}\scriptsize (c) Material classification
	\end{minipage}
	\caption{Visualization of classification results for the MuS-Polar3D objects.}
	\label{fig:classification_results}
\end{figure*}

The resulting cluster assignments are as shown in Fig.~\ref{fig:classification_results}(a) and (b). In terms of material properties, all objects are categorized into three classes: resin sculptures, glazed ceramics, and unglazed ceramics, with representative examples visualized in Fig.~\ref{fig:classification_results}(c). The differences in their reflection mechanisms and the corresponding impact on 3D reconstruction performance have been discussed in Section~\ref{sec:Objects and Viewpoints} .

This section further reports the proportional statistics of the above multi-dimensional categorization to characterize the data distribution. First, object size is quantified using the proportion of valid pixels within the foreground mask, and all samples are grouped into three scale levels (small, medium, and large) via K-means clustering. Specifically, the small-scale category corresponds to an effective pixel ratio range of [0.05, 0.09], accounting for 46\% of the dataset; the medium-scale category spans [0.09, 0.17] with a proportion of 37\%; and the large-scale category covers [0.17, 0.25], representing 17\% of the samples.
Second, texture complexity is quantified using the LBP descriptor, and K-means clustering is applied to divide all samples into three texture levels: dull, balanced, and rich. Samples with dull texture complexity fall within an LBP range of [0.40, 0.58], accounting for 46\% of the dataset; balanced texture complexity corresponds to the range [0.58, 0.87] with a proportion of 32\%; and rich texture complexity spans [0.87, 1.42], representing 22\% of the samples.
Finally, regarding material composition, resin sculptures, glazed ceramics, and unglazed ceramics account for 54\%, 24\%, and 22\% of the target objects in the dataset, respectively.

Based on the above data acquisition and processing pipeline, the MuS-Polar3D dataset is constructed. The dataset contains 42 target objects, each captured under seven scattering levels and four polarization states, with observations from up to five viewpoints. In total, MuS-Polar3D comprises 847 samples, providing a solid data foundation for underwater polarization image analysis and 3D vision research.

\section{Experiments and Results}

Subsequently, polarization-based 3D reconstruction tasks are conducted on the MuS-Polar3D dataset to validate its effectiveness for underwater polarization vision research and to establish an experimental foundation for future studies. In turbid water, a large number of suspended particles induce Rayleigh scattering, which significantly attenuates effective target signals. To address this challenge, inspired by computational imaging principles~\cite{xiang2024computational}, polarization-based 3D reconstruction in scattering media is decoupled into a two-stage processing paradigm, consisting of descattering followed by 3D reconstruction. Specifically, scattering-induced degradations are first suppressed and corrected, and 3D reconstruction is then performed based on the enhanced observations.

For this two-stage pipeline, a systematic evaluation of different algorithms is carried out from both qualitative and quantitative perspectives at each stage, enabling objective performance analysis and the identification of optimal solutions.

\subsection{Experimental Setup}

In the two-stage experimental setting, samples captured under clear water conditions (with 0~ml emulsion added) are first excluded to focus on imaging under scattering environments. The remaining 726 samples are split into training, validation, and test sets with a ratio of 8:1:1. The overall training procedure is organized into two stages: (1) image descattering and (2) 3D reconstruction. 

In the first stage (image descattering), the corresponding clear-water images are used as ground truth to supervise the training of descattering networks. In the second stage (3D reconstruction), ground-truth surface normal maps are employed as supervision to guide the recovery of 3D surface geometry. All experiments are conducted on a computing platform equipped with 4 NVIDIA A100 GPUs, and the training framework is implemented using PyTorch~1.8.0.

\begin{figure*}[!hbp]
	\centering
	\includegraphics[width=0.8\textwidth]{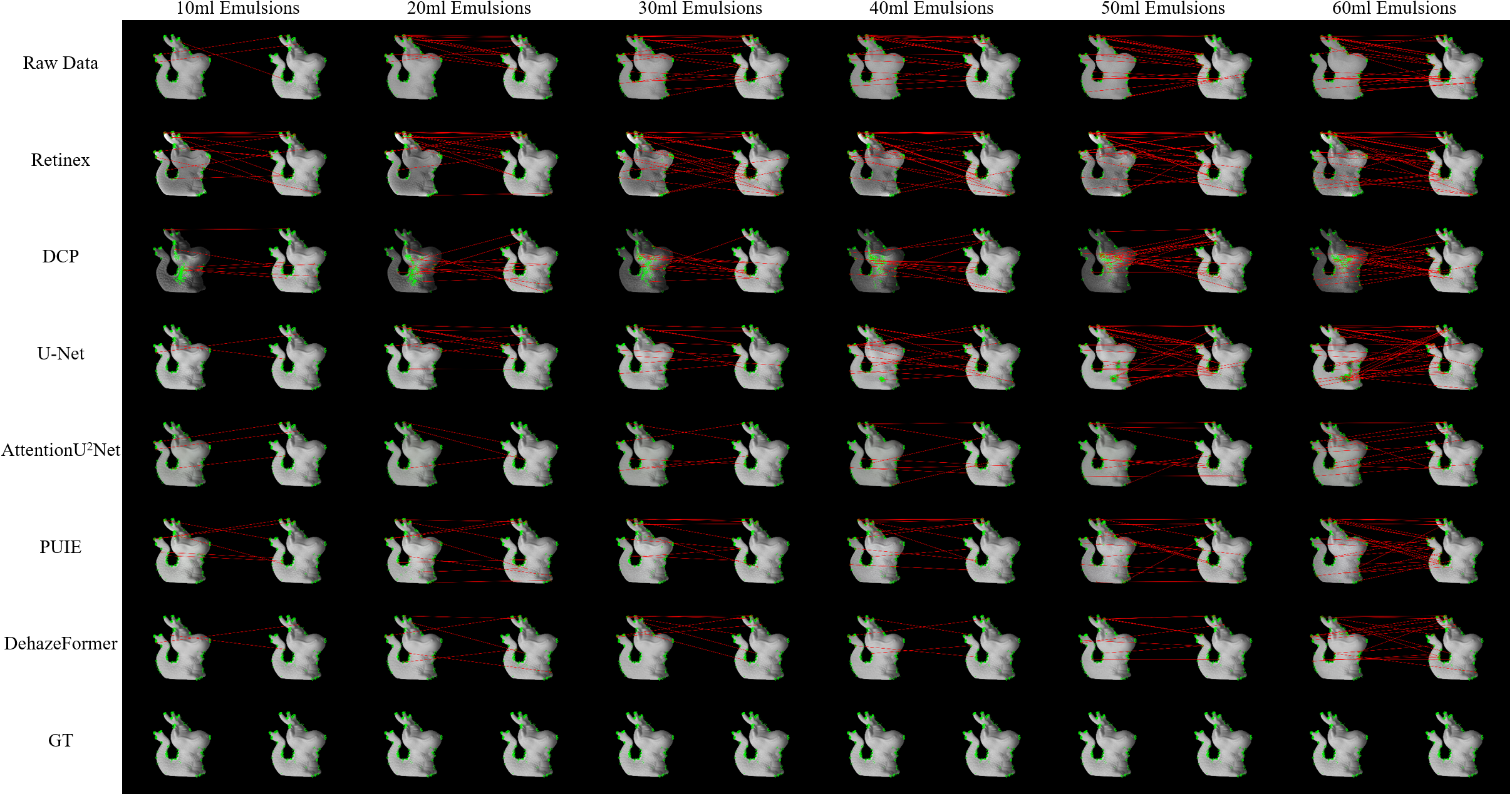}
	\caption{Qualitative analysis of descattering methods based on ORB feature matching. By comparing the number and spatial distribution of feature correspondences between the descattered images and the ground-truth images, the ability of different descattering algorithms to recover geometric structures and local details under varying scattering intensities is evaluated.}
	\label{fig:Qualitative analysis of descattering methods}
\end{figure*}

\subsection{Image Descattering}
\subsubsection{Baselines and Training Strategy}

Six descattering methods are selected for comparison, including Retinex~\cite{fu2014retinex}, Dark Channel Prior (DCP)~\cite{he2010single}, U-Net~\cite{ronneberger2015u}, AttentionU$ ^{2} $Net~\cite{wu2025deep}, PUIE~\cite{fu2022uncertainty}, and DehazeFormer~\cite{song2023vision}. Among them, Retinex and DCP are representative traditional prior-based approaches. U-Net and AttentionU$ ^{2} $Net were originally designed for image segmentation; however, due to the generality of their encoder–decoder architectures, they are adapted for the image descattering task in this work by modifying the output formulation and loss functions.
PUIE and DehazeFormer represent advanced methods in underwater image enhancement and image dehazing, respectively. As their core objective is also to suppress scattering-induced degradation and improve image visibility, they are included as competitive baselines in the comparison experiments.

\begin{figure*}[!hb]
	\centering
	\includegraphics[width=0.8\linewidth]{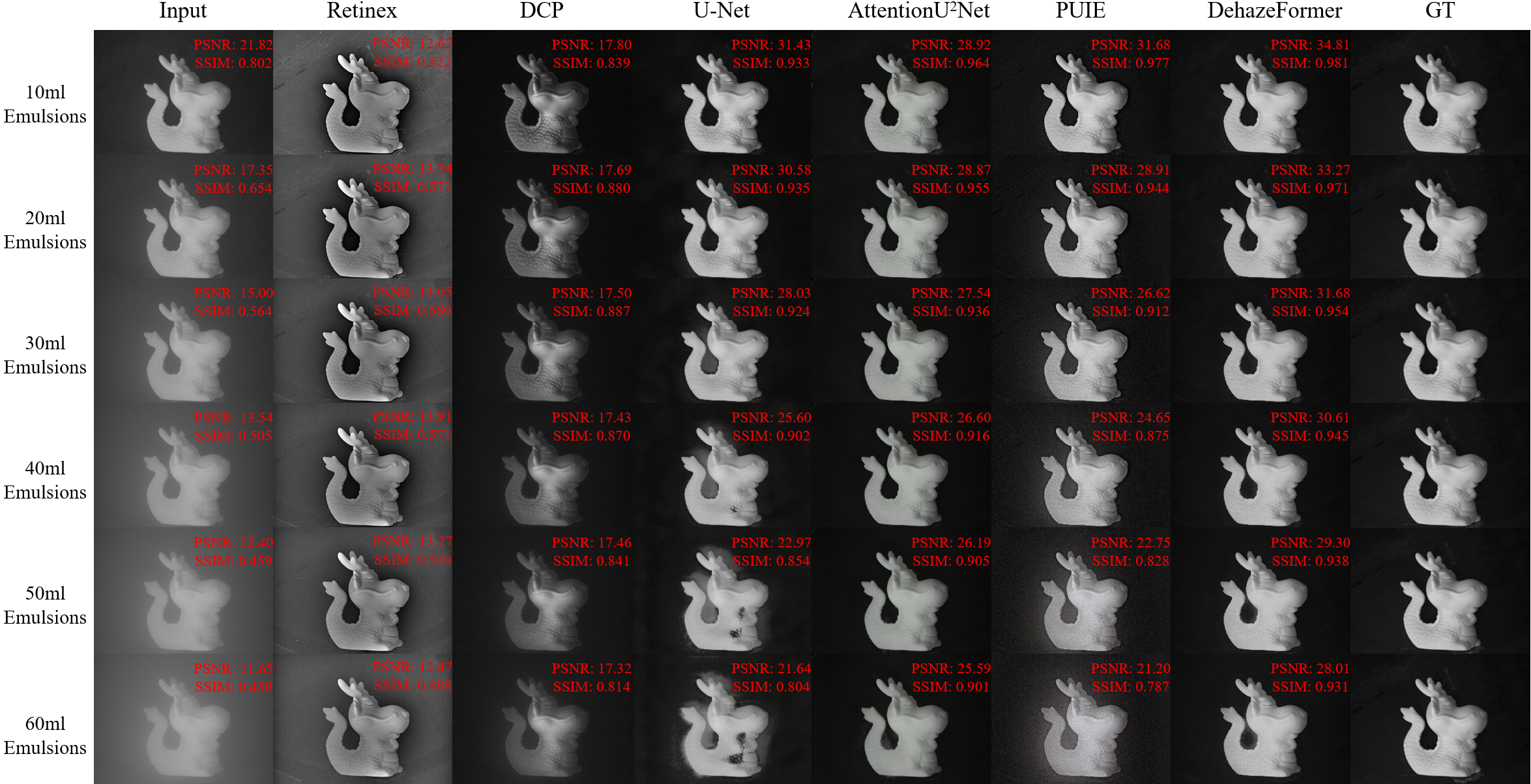}
	\caption{Comparison of descattering results produced by different algorithms for the same target under different scattering levels. The corresponding PSNR and SSIM values are annotated in the upper-right corner of each result image.}
	\label{fig:descattering_results}
\end{figure*}

For U-Net and AttentionU$^{2}$Net, the models are trained from scratch using the Adam optimizer, with a learning rate of 0.001, a batch size of 24, and a total of 100 training epochs. During training, a joint loss function composed of three terms is adopted, which is formulated as
\begin{equation}
	\mathcal{L}_{\mathrm{total}} = 10\mathcal{L}_1 + \mathcal{L}_{\mathrm{SSIM}} + 10\mathcal{L}_{\mathrm{TV}}.
	\label{eq:total_loss}
\end{equation}

The definitions of each loss term are given as follows:
\begin{equation}
	\mathcal{L}_1 = \left\lVert I_{\mathrm{pred}} - I_{\mathrm{gt}} \right\rVert_1,
	\label{eq:l1_loss}
\end{equation}
\begin{equation}
	\mathcal{L}_{\mathrm{SSIM}} = 1 - \mathrm{SSIM}(I_{\mathrm{pred}}, I_{\mathrm{gt}}),
	\label{eq:ssim_loss}
\end{equation}
\begin{equation}
	\mathrm{SSIM}(x,y) =
	\frac{(2\mu_x\mu_y + C_1)(2\sigma_{xy} + C_2)}
	{(\mu_x^2 + \mu_y^2 + C_1)(\sigma_x^2 + \sigma_y^2 + C_2)},
	\label{eq:ssim}
\end{equation}
\begin{equation}
	\begin{aligned}
		\mathcal{L}_{\mathrm{TV}}
		&=
		\frac{1}{N}
		\sum_{x,y}
		\Bigl(
		\lvert I_{\mathrm{pred}}(x,y) - I_{\mathrm{pred}}(x+1,y) \rvert \\
		&\quad +
		\lvert I_{\mathrm{pred}}(x,y) - I_{\mathrm{pred}}(x,y+1) \rvert
		\Bigr).
	\end{aligned}
	\label{eq:tv_loss}
\end{equation}

Here, $I_{\mathrm{pred}}$ and $I_{\mathrm{gt}}$ denote the predicted image and the ground-truth image, respectively, where the ground truth corresponds to images captured under clear-water conditions. The symbols $\mu_x$ and $\mu_y$ represent the mean intensities of image patches $x$ and $y$, while $\sigma_x^2$ and $\sigma_y^2$ denote the corresponding variances, and $\sigma_{xy}$ denotes their covariance. The constants $C_1$ and $C_2$ are introduced to stabilize the structural similarity computation, and $N$ denotes the total number of valid pixels.
In the loss design, $\mathcal{L}_1$ enforces pixel-wise reconstruction accuracy, $\mathcal{L}_{\mathrm{SSIM}}$ emphasizes structural similarity, and $\mathcal{L}_{\mathrm{TV}}$ suppresses noise and artifacts, thereby improving overall image smoothness.

For PUIE and DehazeFormer, fine-tuning is performed based on their officially released pretrained models. Following the training strategies described in the original papers, both models are fine-tuned on the constructed dataset for 100 epochs to fully exploit their performance.

\subsubsection{Qualitative Evaluation}

For image enhancement tasks aimed at improving image visibility, common qualitative evaluation strategies include introducing downstream vision tasks such as object detection~\cite{lei2025yolov13} to compare detection stability and accuracy before and after enhancement. If targets can be detected more stably and accurately in the enhanced images, the enhancement method can be qualitatively considered effective in improving image visibility and detectability. 
Another category of qualitative evaluation relies on upstream vision tasks such as image registration. By analyzing mismatches in feature correspondences between enhanced images and ground-truth images, the enhancement quality can be indirectly assessed from the perspective of feature consistency.

In this study, since background interference in the experimental scenes is relatively limited while the original scattering images commonly suffer from low contrast and blurred details, the ORB feature detector~\cite{rublee2011orb} is adopted to extract keypoints and corresponding descriptors from both the descattered images and the ground-truth images. As illustrated in Fig.~\ref{fig:Qualitative analysis of descattering methods}, under ideal conditions, feature points located at the same spatial positions in the descattered image and the ground-truth image are expected to form one-to-one correspondences.

To this end, the Euclidean distance of each automatically matched feature pair is recomputed in the coordinate systems of the two images. When this distance exceeds a predefined threshold, the match is regarded as an incorrect correspondence and is visualized using a red thin line to enhance intuitive comparison. By examining the number and spatial distribution of incorrect matches, the differences in structural preservation and detail recovery under scattering conditions, as well as across different descattering algorithms, can be directly observed. A smaller number of incorrect matches indicates superior geometric consistency and better recovery of local details after descattering.

As shown in Fig.~\ref{fig:Qualitative analysis of descattering methods}, with the increase in emulsion dosage, the scattering strength of the water gradually intensifies, and the number of incorrect feature matches indicated by red lines exhibits a clear upward trend. This observation suggests that scattering degradation severely disrupts local structural consistency in images, thereby imposing a bottleneck on descattering performance. 

From the qualitative comparison of incorrect match counts, AttentionU$^{2}$Net, PUIE, and DehazeFormer consistently outperform Retinex, DCP, and U-Net on the illustrated samples. Under high-scattering conditions, the former methods are more effective at preserving image structures and fine details, leading to a substantial reduction in incorrect feature correspondences and demonstrating superior descattering capability.

\subsubsection{Quantitative Evaluation}

Four quantitative metrics are adopted to evaluate the performance of different descattering algorithms, including peak signal-to-noise ratio (PSNR), structural similarity index measure (SSIM)~\cite{wang2004image}, learned perceptual image patch similarity (LPIPS)~\cite{zhang2018unreasonable}, and the natural image quality evaluator (NIQE)~\cite{mittal2012making}. LPIPS is computed based on a pretrained VGG model~\cite{simonyan2014very}.

\begin{table}[!t]
	\centering
	\scriptsize
	\setlength{\tabcolsep}{2pt}
	\renewcommand{\arraystretch}{1.05}
	\caption{Quantitative evaluation of descattering algorithms. Red values indicate the best performance among all methods, while blue values denote the second-best results.}
	\begin{tabular*}{\columnwidth}{@{\extracolsep{\fill}} l c cc cccc @{}}
		\toprule
		\multirow{2}{*}{Metric} & 
		\multirow{2}{*}{{Raw Data}} &
		\multicolumn{2}{c}{Traditional} &
		\multicolumn{4}{c}{Learning-Based} \\
		\cmidrule(lr){3-4}\cmidrule(lr){5-8}
		& & Retinex & DCP & U-Net & AttU\textsuperscript{2}Net & PUIE & \makecell{Dehaze\\Former} \\
		\midrule
		PSNR $\uparrow$ & 15.48 & 12.25 & 21.40 & \textcolor{blue}{28.68} & 28.46 & 25.95 & \textcolor{red}{31.51} \\
		SSIM $\uparrow$ & 0.5248 & 0.4286 & 0.8541 & 0.8847 & \textcolor{blue}{0.8948} & 0.8578 & \textcolor{red}{0.9375} \\
		LPIPS$_{\mathrm{VGG}}$ $\downarrow$ & 0.4533 & 0.5310 & \textcolor{red}{0.3669} & 0.5668 & 0.5735 & \textcolor{blue}{0.4035} & 0.4899 \\
		NIQE $\downarrow$ & 19.8191 & 17.1430 & 14.9065 & 13.7683 & 13.0849 & \textcolor{red}{9.2703} & \textcolor{blue}{12.2553} \\
		\bottomrule
	\end{tabular*}
	\label{tab:quantitative_comparison}
\end{table}

As shown in Table~\ref{tab:quantitative_comparison}, deep learning-based descattering methods generally achieve better performance than traditional prior-based approaches across all evaluation metrics. Among them, DehazeFormer stands out by obtaining two best results and one second-best result on the evaluated metrics. It should be noted that none of the tested algorithms are specifically optimized for the LPIPS or NIQE metrics through dedicated loss function design. Under this setting, these two metrics provide a more objective reflection of the overall perceptual quality and naturalness of the descattered results.
Fig.~\ref{fig:descattering_results} further illustrates the descattering results of the six evaluated methods on representative samples. It can be observed that even deep learning-based descattering algorithms may introduce artifacts to varying degrees in regions near the target. Among the evaluated methods, DehazeFormer exhibits the strongest capability in suppressing artifacts, while also achieving the highest PSNR and SSIM scores.

\begin{figure*}[!htbp]
	\centering
	\includegraphics[width=0.8\linewidth]{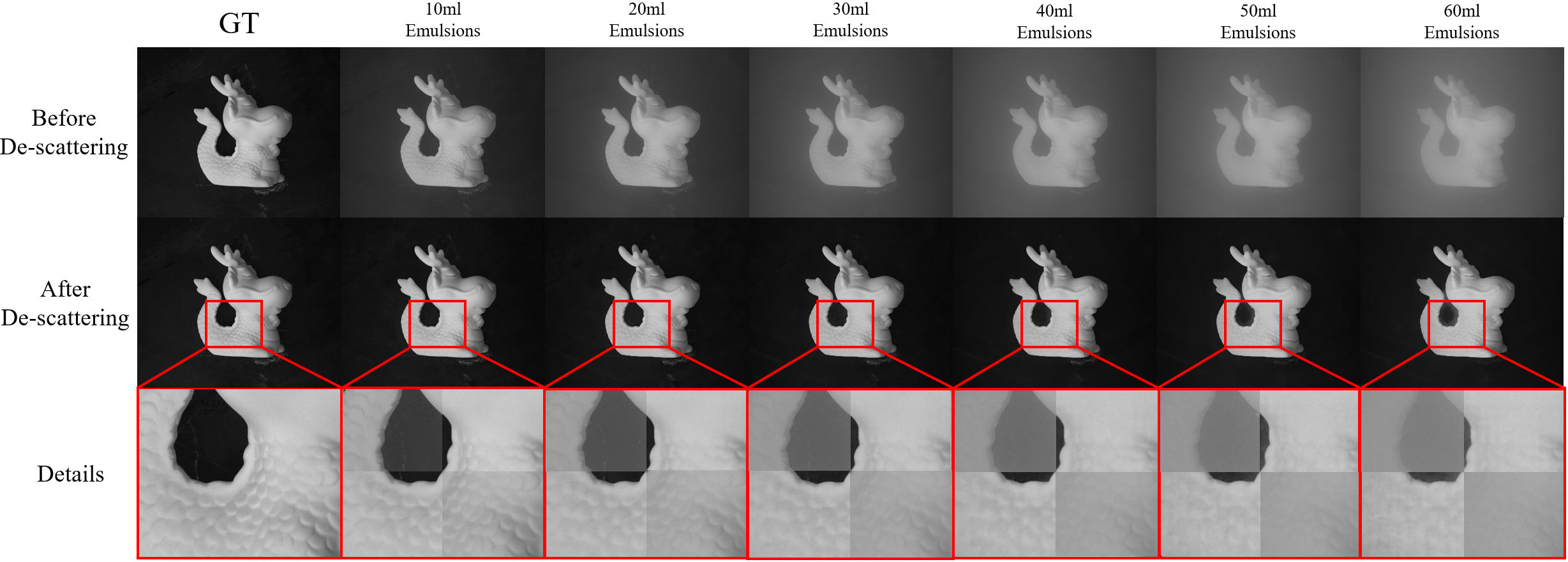}
	\caption{Comparison of descattering results produced by DehazeFormer under different scattering levels. The visual differences before and after descattering are compared, and checkerboard patches are used to highlight local detail differences (red boxes) between the original and descattered images.}
	\label{fig:dehazeformer_visual}
\end{figure*}

\begin{figure}[!htbp]
	\centering
	\begin{minipage}[t]{0.4\linewidth}
		\centering
		\includegraphics[width=\linewidth]{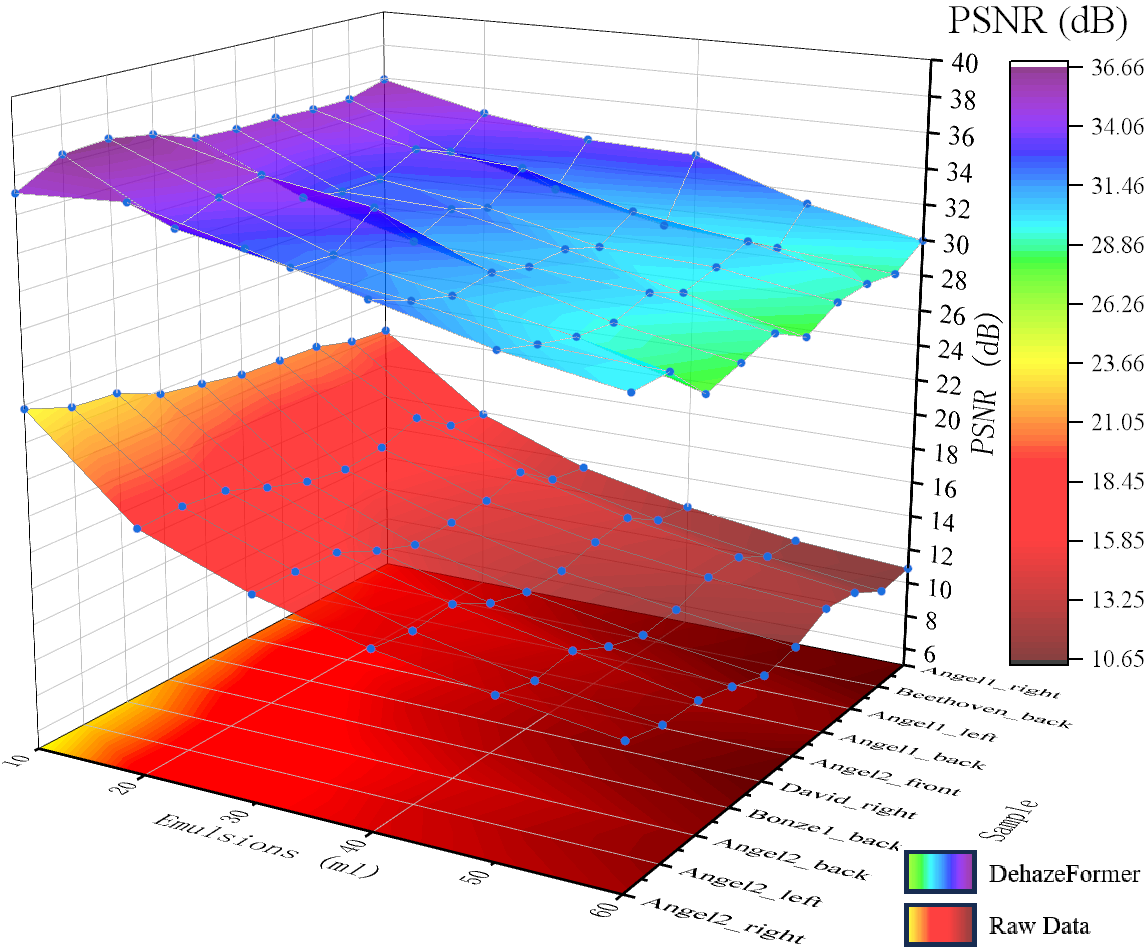} 
		\vspace{2pt}\scriptsize (a) PSNR comparison surface
	\end{minipage}
	\begin{minipage}[t]{0.4\linewidth}
		\centering
		\includegraphics[width=\linewidth]{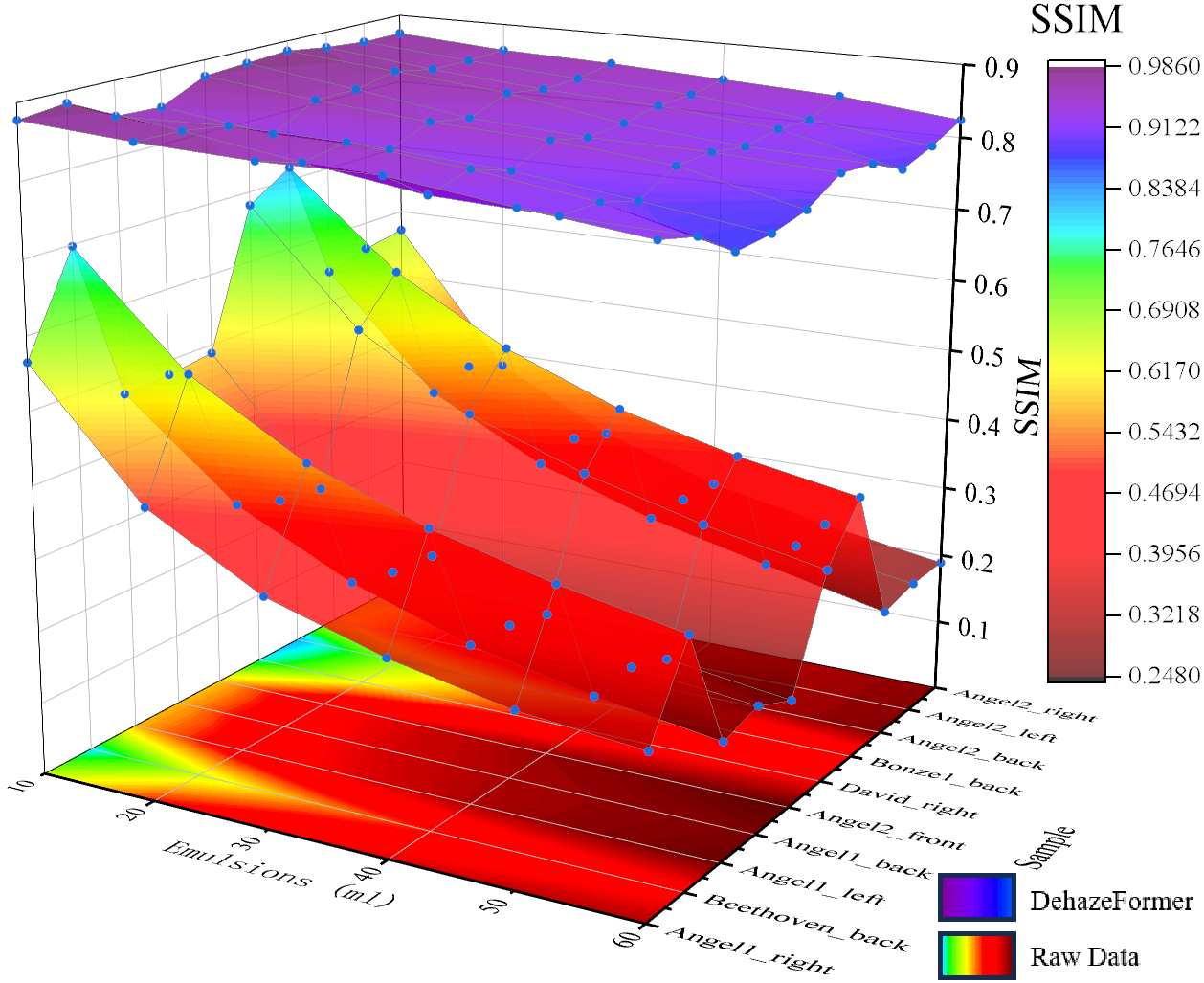} 
		\vspace{2pt}\scriptsize (b) SSIM comparison surface
	\end{minipage}
	\caption{Comparison of PSNR and SSIM trends of the original scattering images and the DehazeFormer-predicted images under different scattering levels.}
	\label{fig:psnr_ssim_trend}
\end{figure}

Given the overall performance advantage of DehazeFormer, its output results are further analyzed in detail. As shown in Fig.~\ref{fig:dehazeformer_visual}, a comparison between images before and after descattering indicates that DehazeFormer demonstrates strong robustness across different scattering levels. Enlarged views of local regions reveal that even under high-scattering conditions, this method can effectively recover object edge structures and fine texture details, thereby significantly improving overall visual quality.
Fig.~\ref{fig:psnr_ssim_trend} presents a visualization of the PSNR and SSIM results for ten representative samples from the test dataset after processing with DehazeFormer. By examining the distribution of the three-dimensional surfaces in Fig.~\ref{fig:psnr_ssim_trend}, it can be observed that as scattering intensity increases, both PSNR and SSIM of the original images and the DehazeFormer results exhibit an overall decreasing trend. However, compared with the original data, the performance degradation of DehazeFormer is noticeably more gradual, further indicating its superior robustness and generalization capability under varying scattering conditions.

Based on the above analysis, DehazeFormer is selected as the primary descattering model to establish an experimental foundation for subsequent investigations into descattering-driven improvements in 3D reconstruction accuracy.

\begin{figure*}[!t]
	\centering
	\begin{minipage}[t]{0.4\textwidth}
		\centering
		\includegraphics[width=\linewidth]{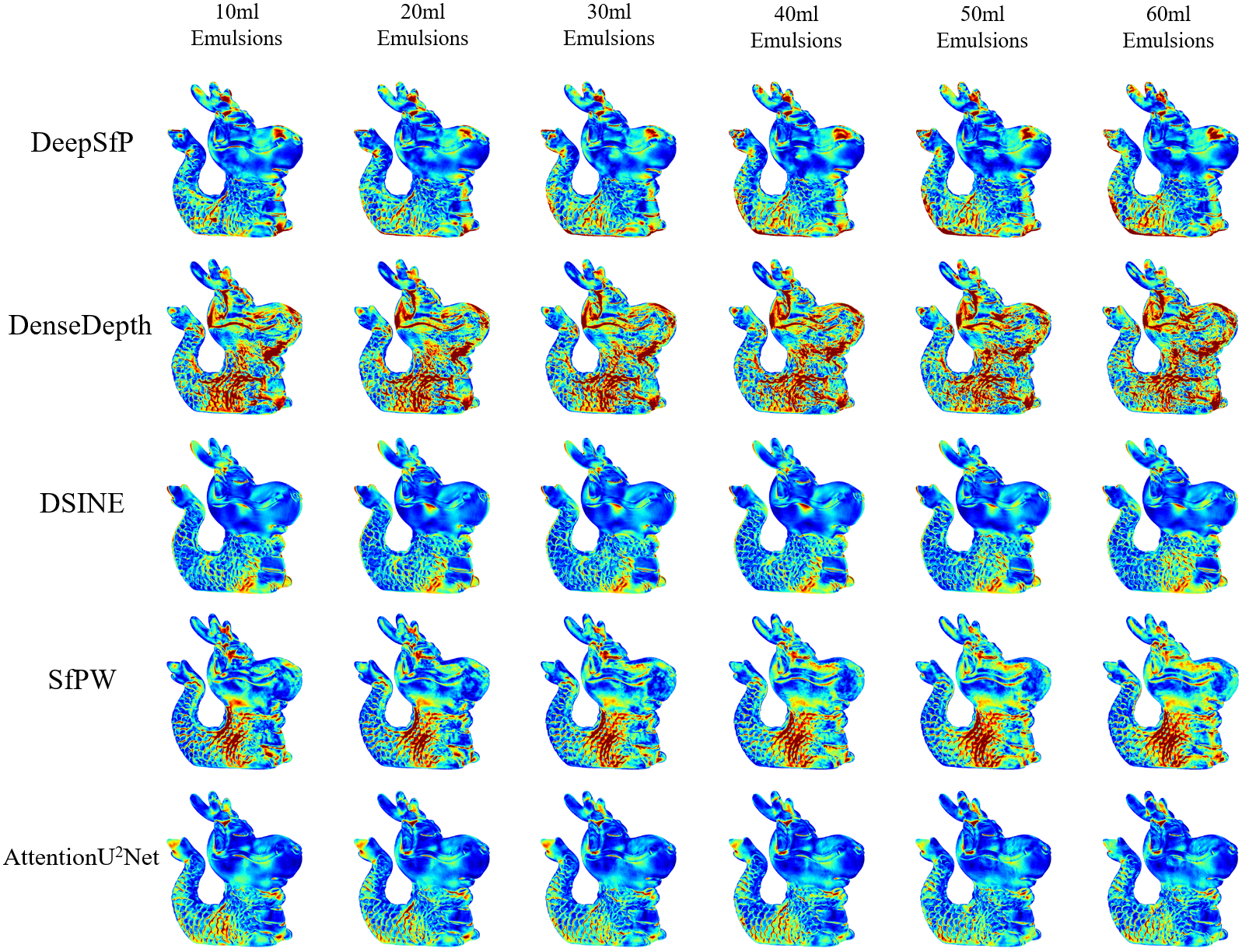}
		\vspace{2pt}
		\scriptsize (a) Error heatmaps \textbf{without} de-scattering.
	\end{minipage}
	\begin{minipage}[t]{0.38\textwidth}
		\centering
		\includegraphics[width=\linewidth]{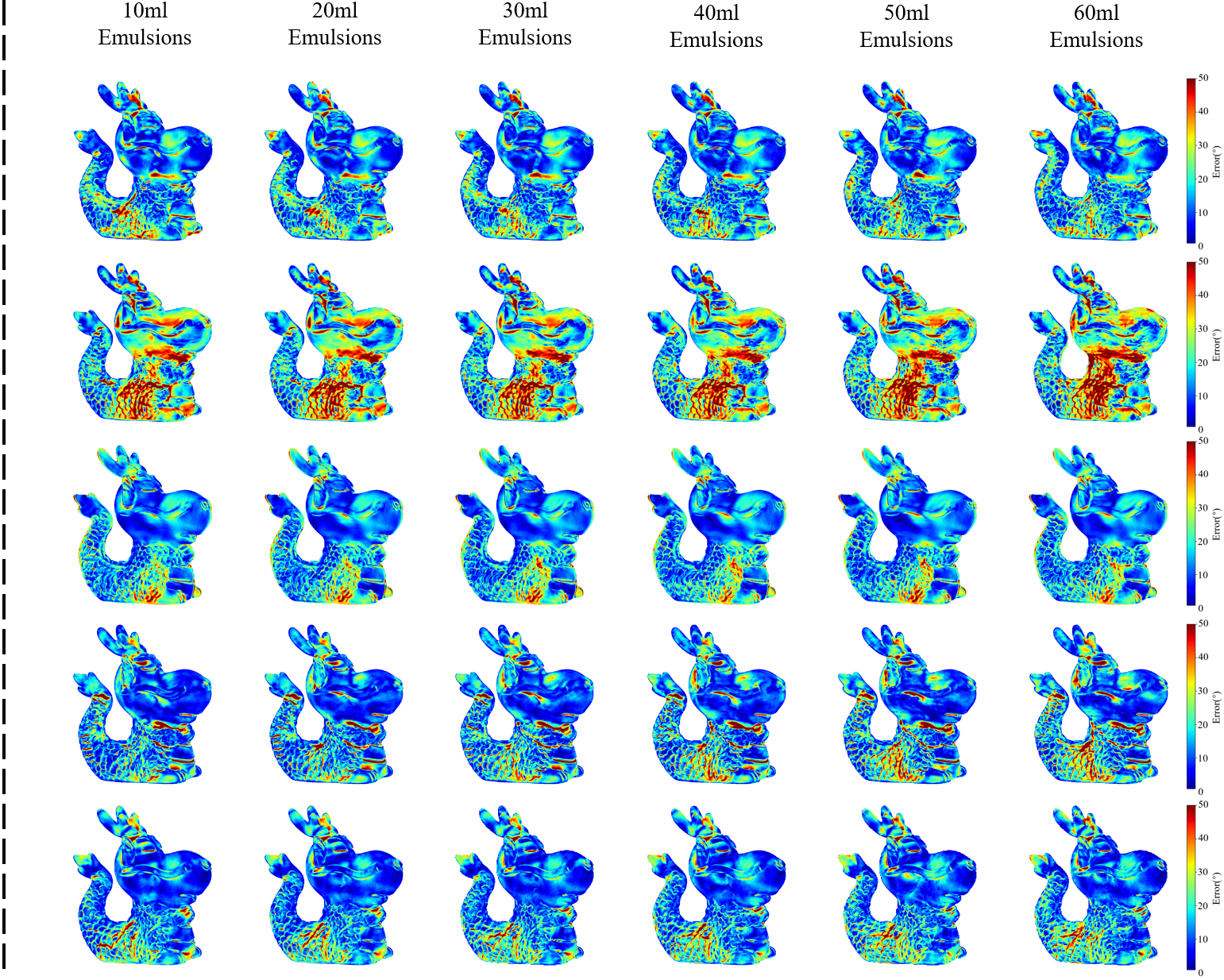}
		\vspace{2pt}
		\scriptsize (b) Error heatmaps \textbf{with} de-scattering.
	\end{minipage}
	
	\caption{Comparison of 3D reconstruction error heatmaps for different algorithms with and without descattering under varying turbidity conditions.}
	\label{fig:heatmap_compare}
\end{figure*}

\subsection{3D Reconstruction}

Building upon the investigation of polarization image descattering, this subsection further explores 3D reconstruction under scattering media conditions. The analysis focuses on the impact of incorporating descattered polarization images as additional inputs to 3D reconstruction networks, compared with using only the original polarization images, on the accuracy of surface normal estimation.

\subsubsection{Baselines and Training Strategy}

Five 3D reconstruction methods are selected for comparison, including the DenseDepth variant from the DSINE repository~\cite{bae2024rethinking}, DSINE~\cite{bae2024rethinking}, DeepSfP~\cite{ba2020deep}, SfPW~\cite{lei2022shape}, and AttentionU$^{2}$Net~\cite{wu2025deep}. Training is conducted on randomly cropped image patches of size $256 \times 256$, with the constraint that the proportion of valid pixels exceeds 50\%.

All methods are trained using the Adam optimizer with a learning rate of 0.001, a batch size of 24, and 300 training epochs. During training, the same loss function is adopted for all methods, defined as
\begin{equation}
	\mathcal{L}_{\mathrm{Cosine}} =
	\frac{
		\sum_{i=1}^{W} \sum_{j=1}^{H}
		\left( 1 - \langle \hat{\mathbf{n}}_{i,j}, \mathbf{n}_{i,j} \rangle \right) - m
	}{
		W \times H - m
	},
	\label{eq:cosine_loss}
\end{equation}
where $W$ and $H$ denote the image width and height, respectively, and $m$ represents the number of invalid pixels excluded by the mask. The vectors $\hat{\mathbf{n}}_{i,j}$ and $\mathbf{n}_{i,j}$ correspond to the predicted surface normal and the ground-truth normal at pixel $(i,j)$. The operator $\langle \cdot , \cdot \rangle$ denotes the cosine similarity between two normal vectors. By minimizing the angular discrepancy between predicted and ground-truth normals, this loss guides the network to learn the spatial geometric structure of local surface elements.

To evaluate the impact of descattering on 3D reconstruction accuracy, two training configurations are considered for comparison: (1) using only the original input to directly predict surface normal vectors of the target objects; and (2) augmenting the original inputs by concatenating the descattered polarization images as additional network inputs during training.

\begin{figure}[!htbp]
	\centering
	\includegraphics[width=0.8\linewidth]{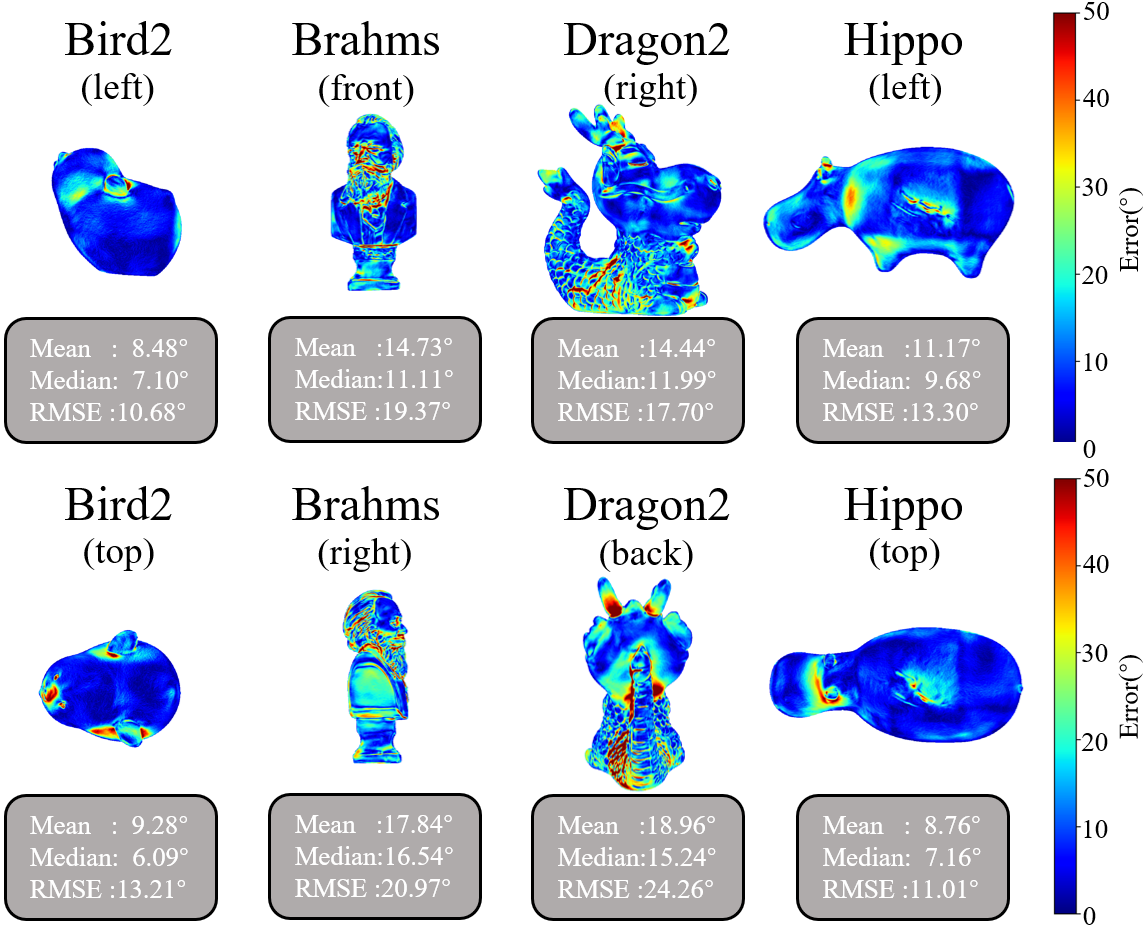}
	\caption{Comparison of error heatmaps for different samples under the same turbidity condition.}
	\label{fig:normal_error_heatmaps}
\end{figure}

\begin{figure*}[!htb]
	\centering
	\includegraphics[width=0.8\linewidth]{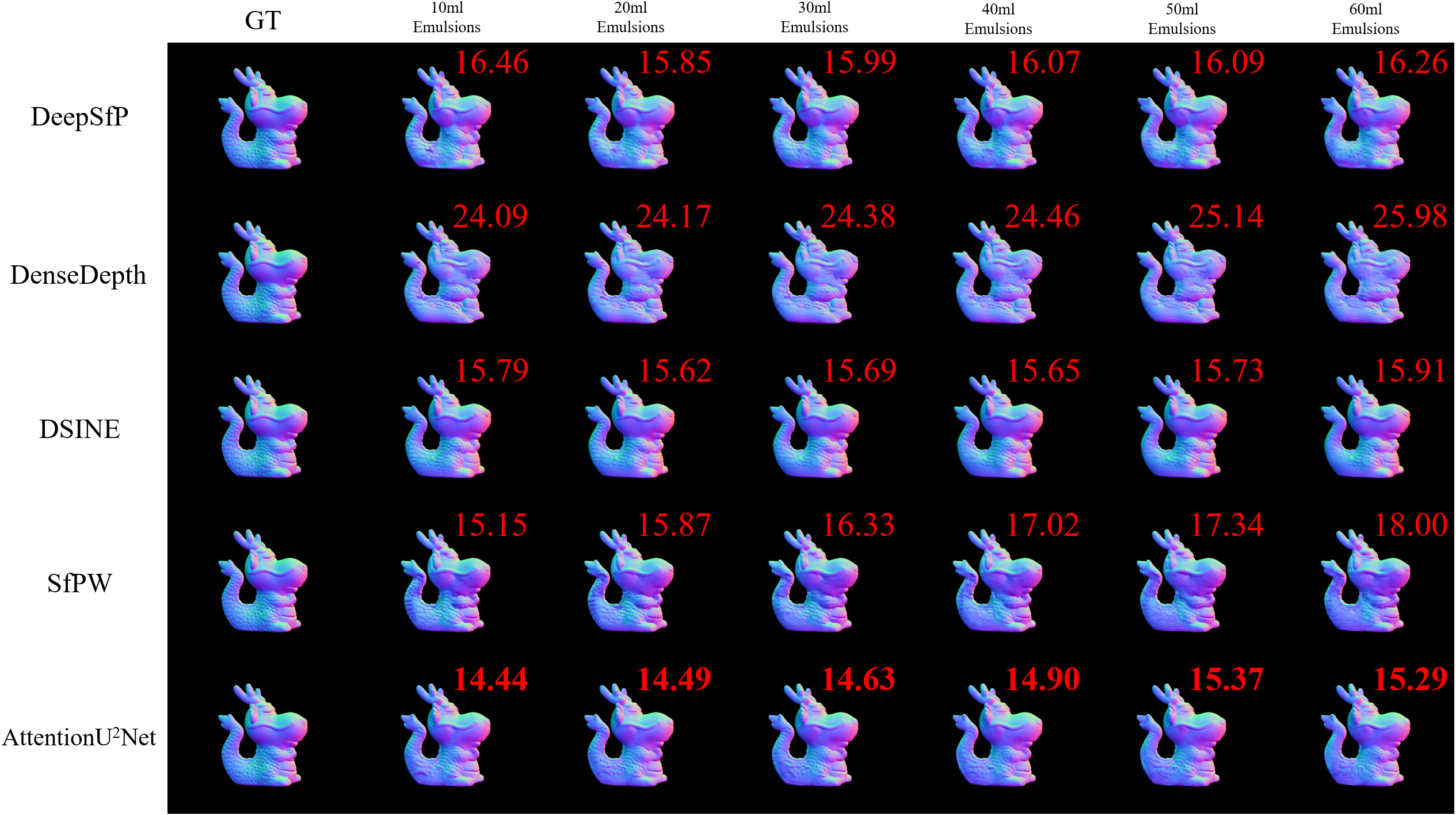}
	\caption{Surface normal prediction results of different baseline methods under different scattering levels after descattering. The red values in the upper-right corner of each image indicate the mean angular error (MAE, in degrees), where lower values correspond to better performance.}
	\label{fig:normal_comparison}
\end{figure*}


\begin{figure*}[!t]
	\centering
	\begin{minipage}[t]{0.5\textwidth}
		\centering
		\includegraphics[width=\linewidth]{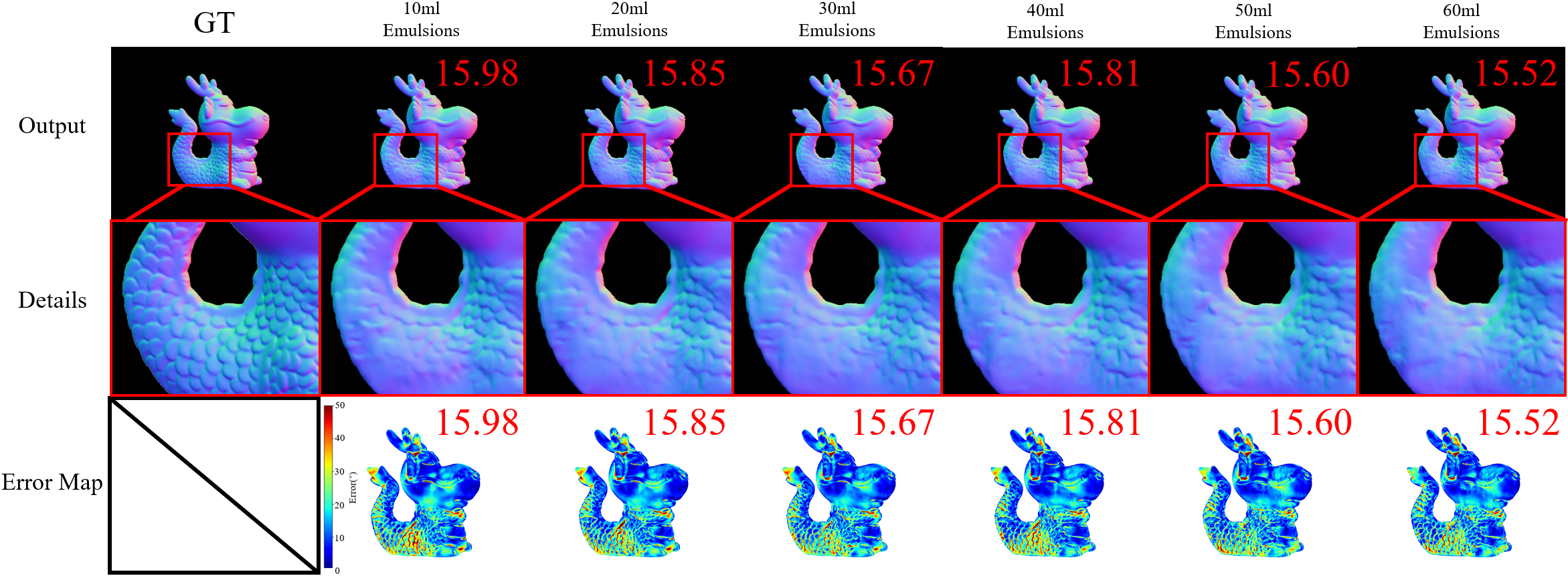}
		\vspace{2pt}
		\scriptsize (a) \textbf{w/o} de-scattering.
	\end{minipage}\hfill
	\begin{minipage}[t]{0.49\textwidth}
		\centering
		\includegraphics[width=\linewidth]{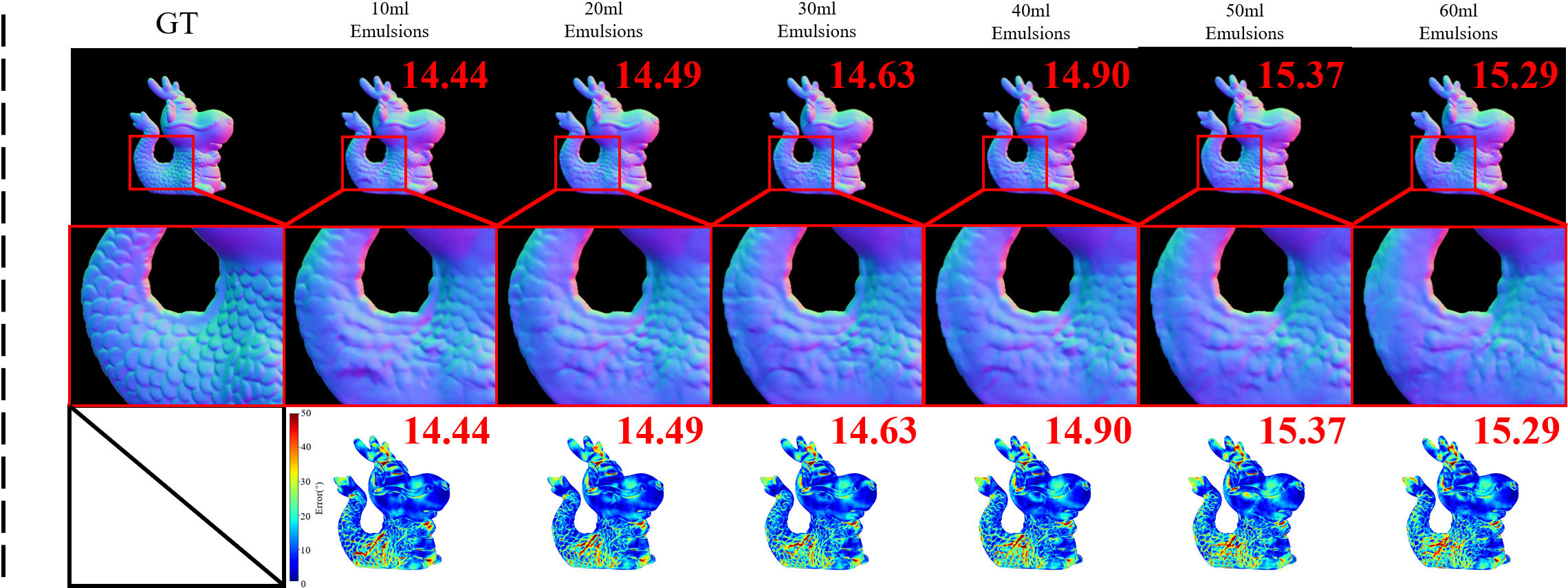}
		\vspace{2pt}
		\scriptsize (b) \textbf{w/} de-scattering.
	\end{minipage}
	
	\caption{Ablation study of AttentionU$^{2}$Net with descattering under different scattering levels. The red values in the upper-right corner of each image indicate the mean angular error (MAE, in degrees), where lower values correspond to better performance.}
	\label{fig:attn_u2net_ablation}
\end{figure*}

\subsubsection{Qualitative Evaluation}

Qualitative error analysis of 3D reconstruction results is conducted by visualizing per-pixel error heatmaps. As shown in Fig.~\ref{fig:heatmap_compare}, the color scale in the heatmaps indicates the magnitude of the error at each pixel, where warmer colors correspond to larger deviations between the predicted and ground-truth surface normals, while cooler colors indicate smaller errors. By examining whether warm or cool colors dominate the heatmaps, the reconstruction quality can be intuitively assessed.
It can be clearly observed from Fig.~\ref{fig:heatmap_compare} that introducing the descattering process leads to a significant reduction in high-error regions and a more spatially smooth error distribution. This improvement is particularly pronounced under high-scattering conditions, highlighting the effectiveness of descattering in enhancing 3D reconstruction performance.

In addition, to further analyze the spatial distribution of reconstruction errors, error heatmaps of different samples are compared under the same turbidity condition and using the same reconstruction algorithm, as illustrated in Fig.~\ref{fig:normal_error_heatmaps}. The results indicate that surface normal errors are mainly concentrated in regions with complex geometry and high-frequency texture patterns, such as the facial region of Brahms and the scale region of Dragon2. In contrast, regions with relatively smooth geometry exhibit noticeably smaller errors, as observed on the torso regions of Bird2 and Hippo.
This phenomenon shows a clear correlation with the texture complexity quantification and categorization results based on the LBP descriptor presented in Section~\ref{sec:data_analysis}, suggesting that texture complexity is a key factor influencing the spatial distribution of 3D reconstruction errors. This correlation not only validates the effectiveness of texture-based quantification in explaining error sources but also provides guidance for developing more targeted improvement strategies for regions with high-frequency textures.

\begin{table}[!h]
	\centering
	\scriptsize
	\setlength{\tabcolsep}{2pt} 
	\renewcommand{\arraystretch}{1.05} 
	\caption{Comparison of the impact of descattering on the performance of different baseline methods on the test dataset. Red values indicate the best results (lower is better).}
	\begin{tabular*}{\columnwidth}{@{\extracolsep{\fill}} lccccc @{}}
		\toprule
		& \multicolumn{5}{c}{Baseline methods} \\
		\cmidrule(lr){2-6}
		& DenseDepth & DSINE & DeepSfP & SfPW & AttU$^2$Net \\
		\midrule
		w/o de-scattering images & 30.20 & 16.94 & 19.64 & 21.64 & \textcolor{red}{15.72} \\
		w/ de-scattering images  & 26.71 & 16.83 & 16.58 & 18.84 & \textcolor{red}{15.49} \\
		Difference        &  3.49 &  0.11 &  3.06 &  2.80 &  0.23 \\
		\bottomrule
	\end{tabular*}
	
	\label{tab:dehaze-baseline}
\end{table}

\begin{table}[!ht]
	\centering
	\scriptsize
	\setlength{\tabcolsep}{3.2pt}
	\renewcommand{\arraystretch}{1.05}
	\caption{Ablation study on the impact of descattering in 3D reconstruction for the representative sample Dragon2. The reported metric is the mean angular error (MAE, in degrees), and red values indicate the best results (lower is better).}
	\begin{tabular*}{\columnwidth}{@{\extracolsep{\fill}} lrrrrrr @{}}
		\toprule
		& \multicolumn{6}{c}{Emulsion volume (ml)} \\
		\cmidrule(lr){2-7}
		Method & 10 & 20 & 30 & 40 & 50 & 60 \\
		\midrule
		\multicolumn{7}{c}{\emph{w/o de-scattering images}} \\
		\midrule
		DeepSfP           & 17.20 & 17.58 & 18.59 & 18.97 & 19.40 & 19.47 \\
		DenseDepth        & 27.07 & 27.37 & 27.66 & 28.21 & 28.28 & 28.75 \\
		DSINE             & \textcolor{red}{15.52} & \textcolor{red}{15.50} & \textcolor{red}{15.54} & \textcolor{red}{15.39} & 15.69 & 16.15 \\
		SfPW              & 20.60 & 20.51 & 21.09 & 21.60 & 22.59 & 23.07 \\
		AttentionU$^2$Net & 15.98 & 15.85 & 15.67 & 15.81 & \textcolor{red}{15.60} & \textcolor{red}{15.52} \\
		\midrule
		\multicolumn{7}{c}{\emph{w/ de-scattering images}} \\
		\midrule
		DeepSfP           & 16.46 & 15.85 & 15.99 & 16.07 & 16.09 & 16.26 \\
		DenseDepth        & 24.09 & 24.17 & 24.38 & 24.46 & 25.14 & 25.98 \\
		DSINE             & 15.79 & 15.62 & 15.69 & 15.65 & 15.73 & 15.91 \\
		SfPW              & 15.15 & 15.87 & 16.33 & 17.02 & 17.34 & 18.00 \\
		AttentionU$^2$Net & \textcolor{red}{14.44} & \textcolor{red}{14.49} & \textcolor{red}{14.63} & \textcolor{red}{14.90} & \textcolor{red}{15.37} & \textcolor{red}{15.29} \\
		\bottomrule
	\end{tabular*}
	
	\label{tab:ablation-Dragon2}
\end{table}

\subsubsection{Quantitative Evaluation}

The mean angular error (MAE) is adopted as the primary quantitative evaluation metric in this subsection. Table~\ref{tab:dehaze-baseline} first reports the impact of introducing descattering on the performance of different baseline methods on the test dataset. It can be observed that after incorporating the descattered polarization images as additional inputs to the 3D reconstruction networks, the reconstruction errors of all baseline methods decrease to varying extents. Among these methods, AttentionU$^{2}$Net achieves the lowest MAE, demonstrating the best overall performance.
Fig.~\ref{fig:normal_comparison} illustrates the visual performance of different baseline methods on representative samples after introducing the descattering process, while Table~\ref{tab:ablation-Dragon2} provides a quantitative comparison of the MAE for the same target under different scattering levels, with and without descattering. Analysis of the results in Table~\ref{tab:ablation-Dragon2} leads to conclusions consistent with the qualitative observations: incorporating descattering consistently improves 3D reconstruction accuracy across varying scattering conditions.

Given that AttentionU$^{2}$Net achieves the best performance among all baseline methods at this stage, its reconstruction results are further analyzed in detail. As shown in Fig.~\ref{fig:attn_u2net_ablation}, with increasing scattering strength, surface normal estimation without descattering gradually degrades in terms of local detail preservation and structural consistency, with errors mainly concentrated around geometric edges and high-curvature regions.

After introducing the descattering process, enlarged views of the representative sample Dragon2 reveal clearer and richer texture structures in the scale regions. In contrast, without descattering, these details are significantly obscured by scattering effects. Moreover, the corresponding error heatmaps show that descattering effectively reduces the spatial extent of high-error regions, indicating an improvement in the accuracy of local geometric recovery.

\begin{figure*}[!htbp]
	\centering
	\includegraphics[width=0.7\linewidth]{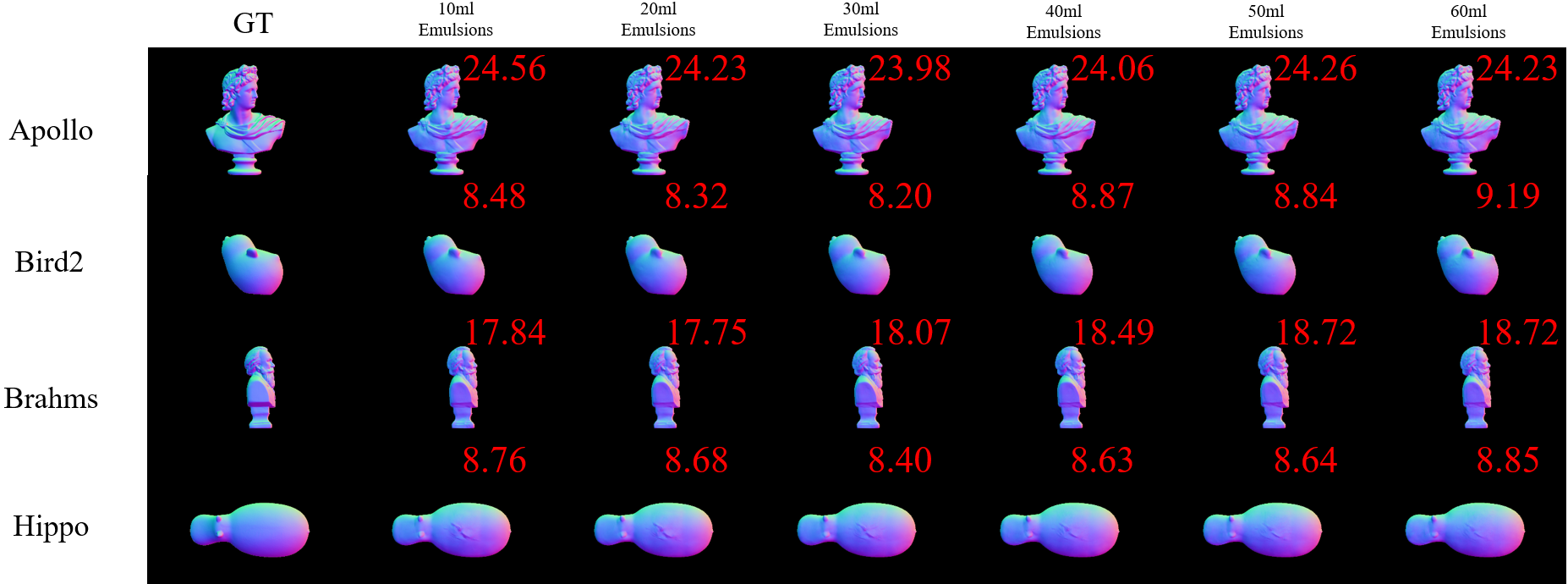}
	\caption{Prediction results of AttentionU$^{2}$Net for different targets under varying scattering levels. The red values in the upper-right corner of each image indicate the mean angular error (MAE, in degrees), where lower values correspond to better performance.}
	\label{fig:object_normal_compare}
\end{figure*}

These results further demonstrate that explicitly modeling and preprocessing scattering degradation provide more stable and higher-quality inputs for downstream tasks, thereby enhancing the reliability of 3D imaging systems in complex underwater environments.
Fig.~\ref{fig:object_normal_compare} presents the 3D reconstruction results of AttentionU$^{2}$Net for multiple samples under different scattering levels after introducing the descattering process. A horizontal comparison indicates that as scattering intensity varies, the mean angular error of each target remains relatively stable without significant performance degradation, demonstrating the robustness and generalization capability of AttentionU$^{2}$Net to changes in scattering conditions.

A vertical comparison reveals that differences in reconstruction error across objects are mainly related to their inherent geometric complexity and the distribution of local high-frequency structures, rather than the scattering strength itself. This observation further suggests that once scattering degradation is effectively suppressed through descattering, the primary limiting factors for 3D reconstruction performance are dominated by the intrinsic geometric properties of the target objects.

\subsection{Further Discussion}

\begin{figure*}[!htb]
	\centering
	\includegraphics[width=0.7\linewidth]{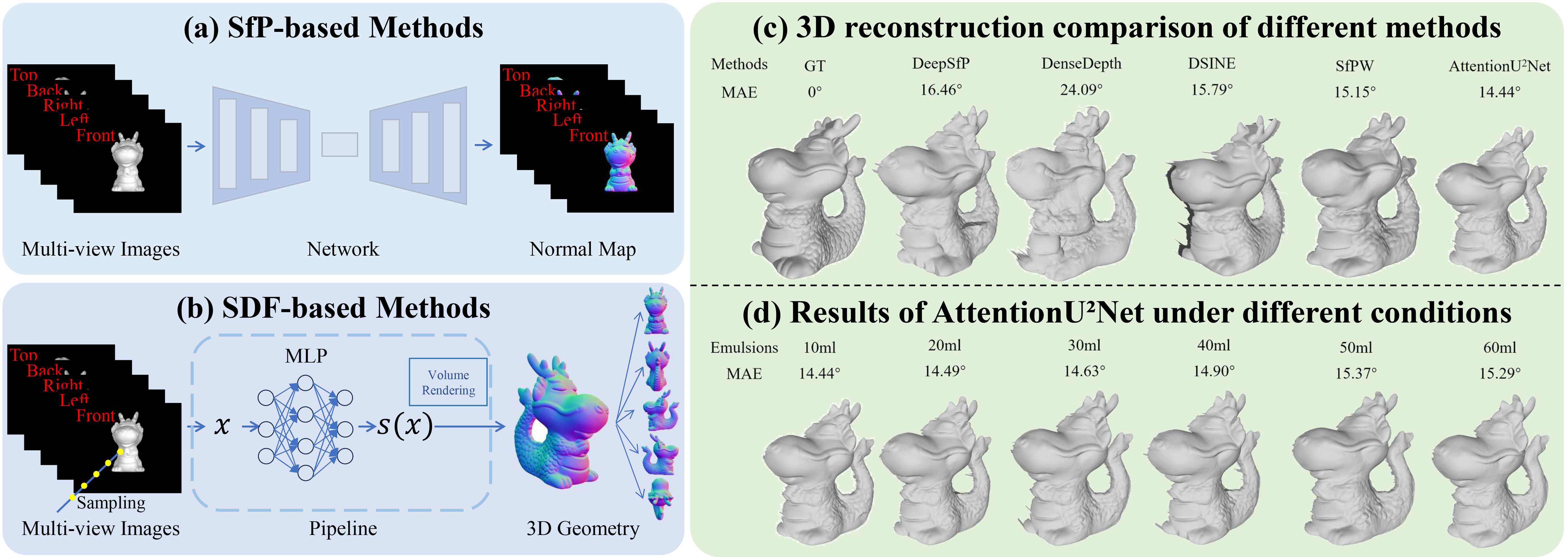}
	\caption{Comparison of normal estimation approaches and resulting 3D reconstructions.}
	\label{fig:3d_reconstruction}
\end{figure*}

Unlike conventional approaches that rely solely on single-view normal prediction (e.g., SfP), the constructed dataset provides consistent multi-view observations under scattering conditions together with high-quality ground-truth annotations. As illustrated in Fig.~\ref{fig:3d_reconstruction}(a) and (b), this property enables single-view normal estimation methods to be compared with SDF-based multi-view normal estimation approaches under a unified data and evaluation framework, which constitutes one of the potential directions for future research.

In addition, to evaluate the impact of normal estimation errors on downstream tasks, the methodology introduced in prior work~\cite{cao2022bilateral} is adopted to integrate the predicted normal maps into complete 3D surface reconstructions. As shown in Fig.~\ref{fig:3d_reconstruction}(c) and (d), comparisons between reconstruction results obtained by different methods under varying scattering levels and the ground-truth integrated surfaces reveal that larger MAE values often lead to noticeable surface flattening or structural distortions in the reconstructed geometry. In contrast, AttentionU$^{2}$Net maintains relatively stable and accurate 3D reconstruction performance across different scattering conditions.

\section{Conclusion}

MuS-Polar3D is an underwater polarization-based 3D imaging dataset that integrates multi-view observations with quantitatively controlled scattering levels. Beyond serving as a benchmark for polarization-based 3D reconstruction in scattering media, MuS-Polar3D bridges the gap between traditional single-view normal estimation methods and SDF-based multi-view normal estimation approaches by providing a unified data and evaluation framework enabled by its multi-view acquisition design. In addition, the dataset exhibits strong extensibility and supports a broader range of vision tasks, such as novel view synthesis, material property estimation, and salient object detection.

Inspired by computational imaging principles, the effectiveness of a two-stage processing paradigm, namely descattering followed by 3D reconstruction, is further validated. Experimental results demonstrate that this strategy consistently improves the recovery of geometric details under varying scattering conditions. Future research will focus on further enhancing high-frequency geometric details and exploring deeper integration of polarization cues with multi-view geometric information.

Overall, MuS-Polar3D provides a general and challenging foundation for advancing integrated research on polarization perception and 3D vision in complex underwater environments.

\bibliographystyle{IEEEtran}
\bibliography{refs}

\vfill

\end{document}